\pdfoutput=1
\documentclass[10pt, logo, twocolumn, copyright, nonumbering]{nvidiatechreport}

\usepackage{hyperref}
\usepackage{cleveref}
\usepackage{url}
\usepackage{xurl}
\usepackage{colortbl}
\usepackage{nicefrac}       %
\usepackage{microtype}      %
\usepackage[dvipsnames]{xcolor}         %
\usepackage{multirow}
\usepackage{multicol}
\usepackage{graphicx}
\usepackage{xspace}
\usepackage{amsmath}
\usepackage{adjustbox}
\usepackage{tcolorbox}
\usepackage{amssymb}
\usepackage{enumitem}
\usepackage{wrapfig}
\usepackage{xcolor}
\usepackage{subcaption}    %
\usepackage{float}
\usepackage{stfloats}
\usepackage{amsmath}
\usepackage{listings}
\usepackage{mdframed}
\usepackage{tcolorbox}
\usepackage{arydshln}
\usepackage{booktabs} 
\usepackage{color,soul}
\usepackage{makecell}
\usepackage[numbers]{natbib}
\usepackage{caption}
\usepackage{makecell}

\tcbuselibrary{listings,breakable}

\definecolor{pastelblue}{RGB}{173,216,230}
\definecolor{pastelyellow}{RGB}{255,253,208}
\definecolor{pastelpink}{RGB}{255,209,220}
\definecolor{pastelgreen}{RGB}{176,226,172}
\definecolor{pastellavender}{RGB}{230,230,250}
\definecolor{TitleCardGreen}{HTML}{EFF7DF}

\definecolor{NvidiaGreen}{RGB}{118, 185, 0}
\sethlcolor{red!15}

\newcommand{\shortname}{Hydra-0\xspace}

\newif\ifshowannotations
\showannotationstrue

\title{\shortname: Action Flow for Generalist World Modeling and Control}

\author{
Hongyu Li\textsuperscript{1,2,3}, Bowen Wen\textsuperscript{1}, Xinghao Zhu\textsuperscript{1}, Yixuan Wang\textsuperscript{1}, Yilun Du\textsuperscript{1,4}, Yunzhu Li\textsuperscript{3} \authorcr George Konidaris\textsuperscript{2}, Stan Birchfield\textsuperscript{1}, Soha Pouya\textsuperscript{1}, Chenran Li\textsuperscript{1}, Yan Chang\textsuperscript{1}\hspace*{\linewidth}\authorcr
\Affilfont\textsuperscript{1}NVIDIA \quad \textsuperscript{2}Brown University \quad \textsuperscript{3}Columbia University \quad \textsuperscript{4}Harvard University
}

\begin{abstract}
We introduce \shortname, a generalist world model conditioned on action flow, which represents robot actions as pixel motion. This shared visual interface enables generalist world modeling and control by learning action consequences across embodiments, tasks, environments, and video-generation backbones. Our best configuration achieves 90.4\% lower robot-motion error and 60.2\% lower object-motion error than our action-conditioned baseline, while supporting zero-shot composition and data-efficient adaptation. On the RoboLab benchmark, \shortname achieves a Pearson correlation of $r=0.96$ between replayed and reference success rates. Finally, we uncover an emergent inverse mode of this interface: a world action model that predicts compatible robot motion from desired object flow transferred from a human demonstration. A trained action head maps the resulting latent features to executable actions without requiring task-specific expert robot demonstrations. Together, these results demonstrate the potential of action flow as a shared control interface connecting heterogeneous training data, open-loop policy evaluation, and robot control.
\end{abstract}

\newcommand{\teaserfigure}{%
    \begin{center}
        \includegraphics[width=0.88\textwidth]{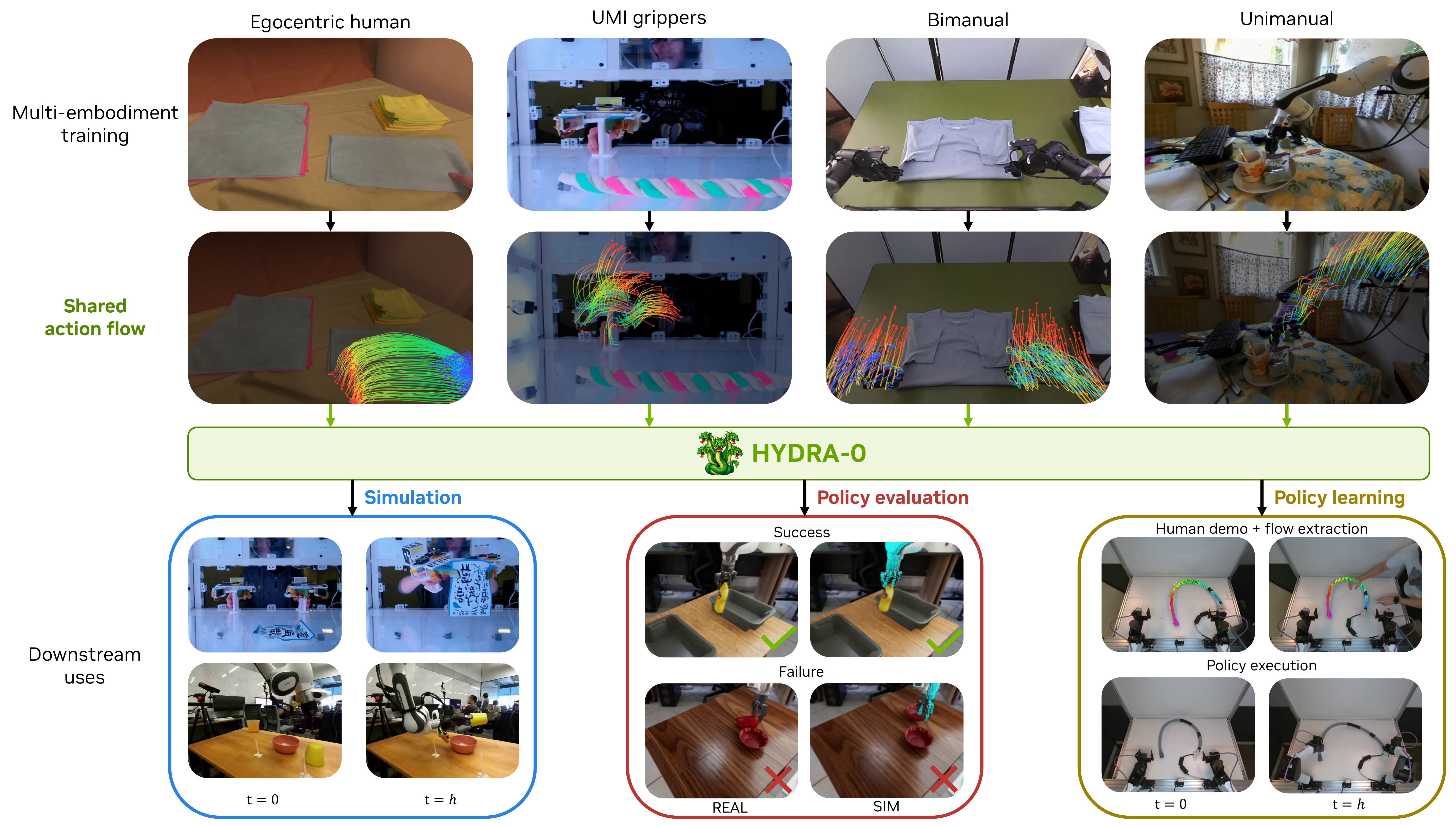}
        \captionof{figure}{\textbf{Action flow as a shared control interface.}
        \emph{Top:} \shortname learns from diverse interaction videos featuring egocentric human demonstrations, handheld UMI grippers, bimanual robot arms, and single-arm robots.
        \emph{Middle:} Visible embodiment motion is represented as image-plane flow trajectories, placing heterogeneous interactions in a common, pixel-aligned action flow space independent of embodiment-specific joint or end-effector coordinates.
        This unified interface enables a single generalist world model to learn from multi-embodiment data and transfer across interaction settings.
        \emph{Bottom:} In forward mode, provided gripper flow conditions future scene prediction, whose rollouts support open-loop policy evaluation; in inverse mode, desired object flow elicits compatible robot motion that a supervised readout converts into executable actions.}
        \label{fig:teaser}
    \end{center}
}

\let\abstractwithoutteaser\abscontent

\fancypagestyle{titlecardstyle}{%
    \fancyhf{}%
    \fancyfoot[L]{\copyrightext}%
    \fancyfoot[R]{\footerfont\thepage}%
}

\makeatletter
\newcommand{\titlecardcontent}{%
    \begin{tcolorbox}[
        colback=TitleCardGreen,
        colframe=TitleCardGreen,
        boxrule=0pt,
        arc=4mm,
        outer arc=4mm,
        boxsep=0pt,
        left=7mm,
        right=7mm,
        top=6mm,
        bottom=2mm
    ]
        {\raggedright\color{black}\sffamily\bfseries\fontsize{21}{24}\selectfont\@title\par}
        \vskip15pt
        {\raggedright\@author\par}
        \vskip13pt
        \abstractwithoutteaser
        \vskip6pt
        {\raggedright\small\sffamily\textbf{Project page:} \url{https://nvidia-isaac.github.io/video_to_data/hydra-0/}\par}
        \vskip4pt
        \hfill\includegraphics[width=78pt,trim=303bp 303bp 303bp 304bp,clip]{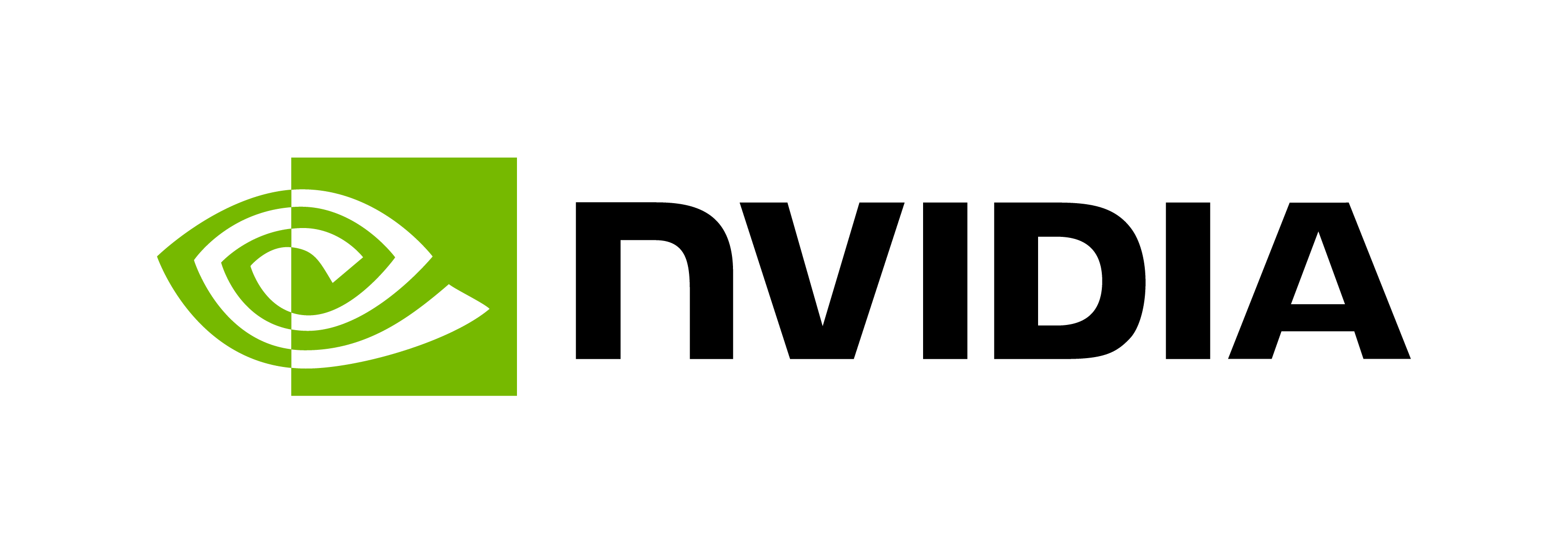}
    \end{tcolorbox}
    \vskip2pt
    \teaserfigure
}
\renewcommand{\maketitle}{%
    \begingroup
    \addtolength{\topmargin}{-30pt}%
    \addtolength{\textheight}{30pt}%
    \onecolumn
    \thispagestyle{titlecardstyle}%
    \titlecardcontent
    \clearpage
    \endgroup
    \twocolumn
}
\makeatother

\begin{document}

\maketitle

\section{Introduction}
Robots that act in the physical world can benefit from predicting how a scene or object will evolve under their motions, a capability provided by a \emph{world model}.
A foundation world model trained on diverse interaction data could serve as a common predictive simulator for many robots and interaction settings by capturing the physical regularities shared across them.
Despite the breadth of available data~\citep{walke_BridgeData_2023,khazatsky_DROID_2024,agibot-world-contributors_AgiBot_2025,li_deform360_2026,allshire_scalable_2026,fang_molmoact2_2026,hoque_EgoDex_2025}, we still lack transferable foundation world models that generalize across embodiments, tasks, and environments.

Many existing world models and action-conditioned video models remain tied to their training embodiment, limiting multi-embodiment training and transfer to new deployment settings~\citep{wang_Interactive_2026,nvidia_World_2026,guo_CtrlWorld_2025,zhou_DINOWM_2025,wang_temporal_2026,wang_adajepa_2026}.
A common reason is their reliance on native robot commands: joint-space commands directly encode a robot's structure, while the same end-effector command can produce different joint trajectories and visible link motions on robots with different kinematics.
Because neither representation specifies the resulting image-plane motion, a video model must learn an embodiment-dependent mapping from commands to visual dynamics.
Motion- and trajectory-conditioned video models demonstrate that image-plane motion can instead provide a shared visual condition~\citep{geng_Motion_2024,wang_ATI_2025a,chu_WanMove_2025}, while flow- and trajectory-based interfaces have also been explored for robot learning and control~\citep{bharadhwaj_Track2Act_2024,xu_Flow_2025,li_NovaFlow_2025,fu_NovaPlan_2026}.
Together, these lines of work leave open how to ground a shared visual motion representation in executable robot commands across embodiments.

We bridge this gap with \emph{action flow}, a shared image-plane motion representation that supports both forward dynamics prediction and inverse motion prediction.
Action flow represents sparse trajectories of visible robot or object points, allowing the same format to describe embodiment motion and task intent.
In the forward direction, we execute a candidate motor command through the robot controller and physics simulation~\citep{nvidia_isaac_2025}, then use robot geometry and camera calibration to project the resulting visible robot-surface trajectories into the image plane.
These pixel-aligned trajectories specify where the robot should move in the observed video while retaining the constraints imposed by its kinematics, providing a directly visual condition tied to an executable command.
Building on trajectory-based visual conditioning in ATI and Wan-Move~\citep{wang_ATI_2025a,chu_WanMove_2025}, our world model uses this condition to predict the command's consequences, a mode we call \emph{kinematically grounded video prediction}.

The same action-flow interface can also be inverted for robot control.
We instantiate this inverse mode as a world action model.
Given desired object flow as task intent, the model infers compatible robot motion, and a target-embodiment readout decodes its latent motion features into executable actions.
We train the model using paired real-world rollouts, including both successes and failures, to infer the robot motion that produces the observed object motion and to recover the corresponding executed actions through the readout.
Together, the forward and inverse modes make action flow a common interface for transfer, simulation, and control.
By expressing motion in the coordinates of the visual observation, action flow allows heterogeneous interaction data to train a generalist predictive model that transfers across embodiments.
\emph{Rather than reproducing task-specific behavior, \shortname models the consequences of robot motion.}

We instantiate our model on different video-generation backbones~\citep{nvidia_World_2026,wan_Wan_2025}, demonstrating that action flow is a portable conditioning interface rather than a backbone-specific modification.
Our best configuration achieves 90.40\% lower robot-motion error and 60.16\% lower object-motion error than our action-conditioned Cosmos 2.5 baseline.
We further evaluate transfer to robot interaction with deformable objects, including zero-shot composition of robot-action grounding with human-demonstrated deformation dynamics and data-efficient task adaptation.
Beyond prediction, our open-loop policy evaluation covers 5 RoboLab policies~\citep{yang_RoboLab_2026}; replayed versus reference success rates have a Pearson correlation of $r=0.96$ across those aggregates.
Finally, we demonstrate real-robot control by converting object flow transferred from a human demonstration into executable actions.

\paragraph{Contributions:}
\begin{enumerate}
    \item We formulate robot action conditioning as kinematically grounded image-plane action flow.
    During training, action flow can be recovered directly from interaction videos; for forward prediction at deployment, we derive it from executable commands using controller-and-physics rollouts in Isaac Lab, robot geometry, and camera calibration.
    \item We show that action flow can serve as a shared interface for generalist world modeling across embodiments and video-generation backbones~\citep{nvidia_World_2026,wan_Wan_2025}.
    Our best configuration achieves 90.40\% lower robot-motion error and 60.16\% lower object-motion error than our action-conditioned Cosmos 2.5 baseline and supports zero-shot and data-efficient transfer to robot--deformable-object interaction.
    \item We use action flow-conditioned simulation for open-loop policy evaluation.
    Across 5 RoboLab policies, replayed versus reference success rates have a Pearson correlation of $r=0.96$.
    \item We invert the same interface in a proof-of-concept world-action model that predicts compatible robot motion from desired object flow transferred from a human demonstration; a trained action head maps its latent features to executable real-robot actions without task-specific expert robot demonstrations.
\end{enumerate}

\definecolor{flowEmb}{RGB}{255,96,48}
\definecolor{flowObj}{RGB}{72,224,136}

\begin{figure*}[t!]
    \centering
    \includegraphics[width=\textwidth]{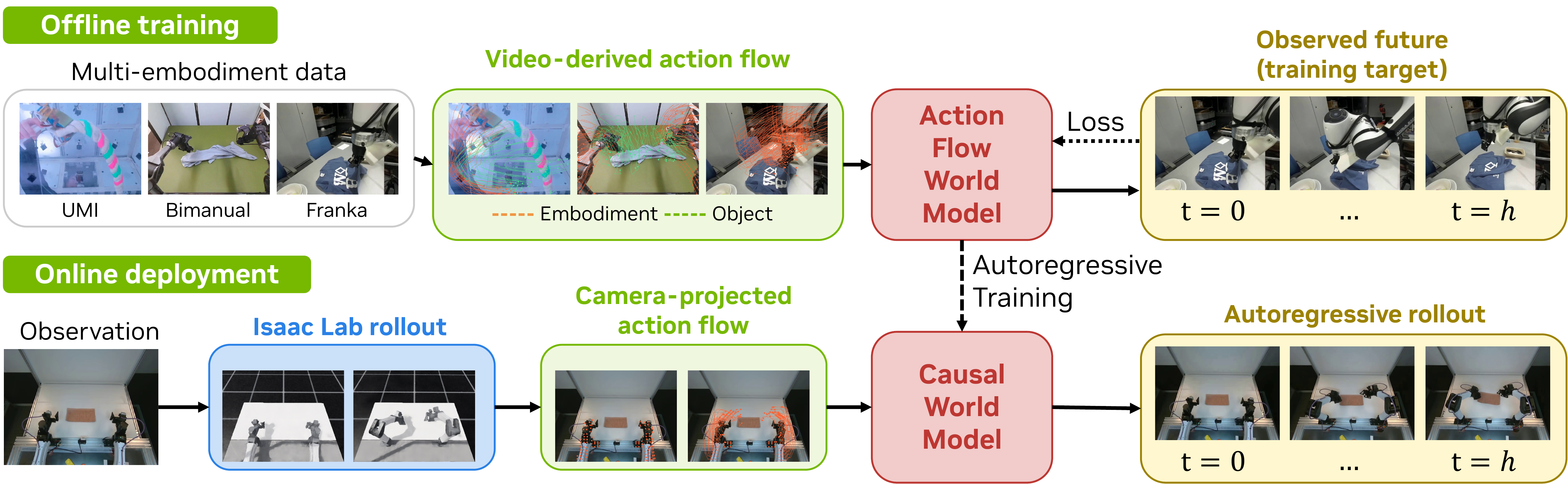}
    \caption{\textbf{Overview of \shortname.}
    During \emph{offline training (top)}, we recover action flow from interaction videos by tracking both the \textcolor{flowEmb}{\textbf{embodiment}} and the manipulated \textcolor{flowObj}{\textbf{object}}, as detailed in \Cref{sec:action_flow}.
    The future video is encoded to construct the flow-matching training target.
    During \emph{online deployment (bottom)}, Isaac Lab executes a candidate command sequence $\mathbf{a}_{0:H-1}$ through the robot controller and physics simulation, producing robot link transforms over the prediction horizon.
    Using these transforms, we propagate visible robot-surface points and project them through the calibrated camera following \Cref{eq:kinematic_projection}, producing embodiment action flow that conditions the causal autoregressive model to predict the command sequence's real-world consequence.
    }
    \label{fig:method_overview}
\end{figure*}

\section{Method}
\label{sec:method}

We consider a partially observable setting in which the robot receives an RGB observation $\mathbf{o}_t\in\mathcal{O}$ and executes an action $\mathbf{a}_t\in\mathcal{A}$ at time $t$.
An encoder $e_\phi$ maps the initial observation to a spatial latent state $\mathbf{s}_0=e_\phi(\mathbf{o}_0)\in\mathcal{S}$.
We represent the action sequence by action flow $\mathcal{F}$, defined below, and let the dynamics model predict the future latent sequence as $\hat{\mathbf{s}}_{1:H}=g_\theta(\mathbf{s}_0,\mathcal{F})$.
A decoder produces the corresponding RGB sequence, $\hat{\mathbf{o}}_{1:H}=d_\psi(\hat{\mathbf{s}}_{1:H})$.
\Cref{fig:method_overview} provides an overview of the resulting world model.

\subsection{Constructing and Sampling Action Flow}
\label{sec:action_flow}

Rather than conditioning the model directly on an embodiment-dependent command, we map the command to \emph{action flow}, a set of camera-plane trajectories describing the commanded motion of the \emph{visible} robot embodiment.
For $N$ tracked points over a prediction horizon of $H$ future steps, we write the image-plane location of point $n$ at time $t\in\{0,\ldots,H\}$ as $\mathbf{x}_{n,t}=(u_{n,t},v_{n,t})$ and its visibility as $m_{n,t}\in\{0,1\}$, so the shared trajectory condition is $\mathcal{F}=\{\boldsymbol{\tau}_n\}_{n=1}^{N}$ with $\boldsymbol{\tau}_n=\{(\mathbf{x}_{n,t},m_{n,t})\}_{t=0}^{H}$.
Because $\mathcal{F}$ is expressed in the observation plane, the same interface can represent robot arms, grippers, and even human hands without exposing their native action spaces to the video model.
We consider two complementary construction scenarios: one in which robot geometry and camera calibration are available, and one in which they are unavailable.
Accordingly, we describe the geometry-aware and video-only construction routes in parallel below.

\paragraph{Geometry-aware construction.}
When robot geometry and camera calibration are available, we sample points on the robot surface visible in $\mathbf{o}_0$ and obtain a sequence of robot states and link transforms.
At deployment, Isaac Lab produces this sequence by executing the candidate command through the robot controller and physics simulation.
We propagate the sampled surface points using these link transforms and project the resulting motion into the camera plane.
For tracked point $n$, let $\mathbf{X}_n$ denote its 3D position in the coordinate frame of robot link $\ell(n)$.
Its image position at time $t$ is
\begin{equation}
    \mathbf{x}_{n,t}
    =\pi\!\left(\mathbf{K}\,\begin{bmatrix}\mathbf{I}_3 & \mathbf{0}\end{bmatrix}\mathbf{T}_{CW}\,\mathbf{T}_{\ell(n)}(\mathbf{q}_t)\,\bar{\mathbf{X}}_n\right),
    \label{eq:kinematic_projection}
\end{equation}
where $\mathbf{q}_t$ is the robot configuration at time $t$, produced by the controller-and-physics rollout in Isaac Lab at deployment or obtained from recorded robot states.
$\mathbf{T}_{\ell(n)}$ is the corresponding link transform.
$\mathbf{T}_{CW}$ and $\mathbf{K}$ are the camera extrinsics and intrinsics, respectively.
$\begin{bmatrix}\mathbf{I}_3 & \mathbf{0}\end{bmatrix}$ selects the Euclidean camera coordinates from the homogeneous vector.
$\bar{\mathbf{X}}_n$ is the homogeneous surface point, and $\pi$ performs perspective projection.
We set $m_{n,t}=1$ only when the projected point has positive camera depth, lies within the image bounds, and agrees with the rendered depth buffer within a small tolerance; a $3\times3$ neighborhood avoids false rejection at silhouette boundaries.

The projected visible trajectories form the deployment action flow.
Moderate calibration error perturbs the two-dimensional motion condition without altering the target video, whereas large projection errors can degrade spatial correspondence.

\paragraph{Video-only construction.}
Many large interaction datasets provide videos but omit robot description files or camera calibration.
For these data, we first recover dense image-plane trajectories and visibility labels with a flow tracker~\citep{harley_AllTracker_2025}, then segment the tracks using grounded masks for the visible embodiment and manipulated objects~\cite{carion_SAM_2025}.
This produces the same trajectory representation $\mathcal{F}$ for training without requiring privileged metadata.
The tracked future is used only to construct training conditions; at deployment, robot action flow is computed causally from the corresponding physics simulator command rollout.

\begin{figure}[t!]
    \centering
    \includegraphics[width=\columnwidth]{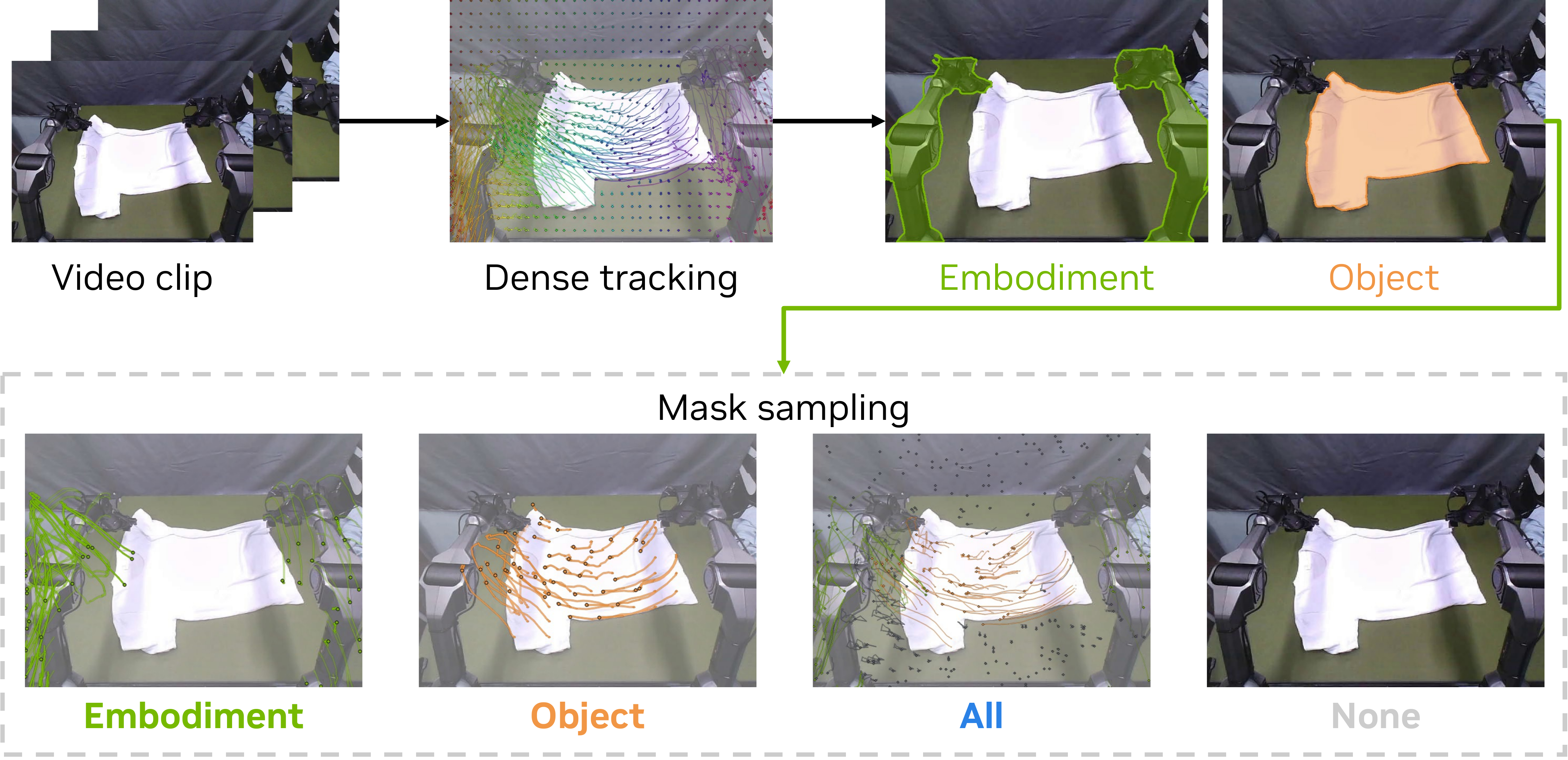}
    \caption{\textbf{Action Flow Construction and Sampling.}
    Dataset videos without geometric information are tracked densely and grounded into embodiment and manipulated-object masks.
    During training, we sample an input condition from four conditioning modes: \emph{Embodiment} selects acting-body tracks as the primary action condition; \emph{Object} supplies task or desired-motion flow; \emph{All} provides a grounding fallback; and \emph{None} provides conditioning dropout.}
    \label{fig:action_flow_sampling}
\end{figure}

\Cref{fig:action_flow_sampling} illustrates this video-only route and the four training-time sampling strategies described next; the geometry-aware route appears in the deployment lane of \Cref{fig:method_overview}.

\paragraph{Training-time flow sampling.}
During training, we sample multiple flow conditions rather than relying on a single track category.
This also enables the model to use different motion cues across applications~\cite{liu_MoRight_2026, gillman_goal_2026}.
At each training step, we sample one of four conditioning strategies from the grounded track set: \emph{Embodiment}, \emph{Object}, \emph{All}, or \emph{None}.
Instead of generating a separate learned token, the sampled mode only determines which trajectories populate the shared motion tensor.
\emph{Embodiment} selects tracks on the acting body and is our primary action-conditioning strategy; it includes visible robot links and grippers or human hands.
\emph{Object} selects observed future tracks on the manipulated object during training; when an object trajectory is explicitly supplied at inference~\citep{li_NovaFlow_2025,fu_NovaPlan_2026}, the same mode can instead provide desired-motion conditioning that elicits compatible robot motion through the world model.
\emph{All} samples across grounded and unassigned tracks so that training retains approximate motion correspondence when semantic grounding is incomplete.
\emph{None} removes the trajectory condition and acts as conditioning dropout, so the model predicts based only on text and image conditioning.
If the tracks required by a strategy are unavailable, we sample from the remaining valid strategies.

\subsection{Action-Flow-Conditioned Video Prediction}

Over a horizon of $H$ steps, our kinematically grounded video predictor produces future observations conditioned on action flow as
\begin{equation}
    \hat{\mathbf{o}}_{1:H}=d_\psi\!\left(g_\theta\!\left(e_\phi(\mathbf{o}_0),\mathcal{F}\right)\right).
    \label{eq:flow_conditioned_prediction}
\end{equation}
For motion conditioning, we retain the spatial structure of the initial encoded state $\mathbf{s}_0$.
Inspired by ATI~\citep{wang_ATI_2025a} and Wan-Move~\citep{chu_WanMove_2025}, we align each image-space trajectory with the temporally compressed video latent by pooling its positions and visibility over the image frames represented by each latent frame.
We express the resulting trajectory locations and destination cells in a common normalized latent-grid coordinate system, denoting the pooled location and visibility at latent time $k$ by $\widetilde{\mathbf{x}}_{n,k}$ and $\widetilde{m}_{n,k}$, respectively.
For each selected trajectory, we then bilinearly sample a source feature $\mathbf{h}_n=\mathbf{s}_0(\widetilde{\mathbf{x}}_{n,0})$ and \emph{propagate it along the temporal dimension} around its future latent-grid locations using Gaussian weights.
At latent time $k$ and normalized latent-grid location $\widetilde{\mathbf{p}}$, the motion feature is
\begin{equation}
    \begin{aligned}
        M_k(\widetilde{\mathbf{p}})
        &=\sum_{n\in\mathcal{N}_K(\widetilde{\mathbf{p}},k)}
        \widetilde{w}_{n,k}(\widetilde{\mathbf{p}})\,\mathbf{h}_n, \\
        \widetilde{w}_{n,k}(\widetilde{\mathbf{p}})
        &=\widetilde{m}_{n,0}\widetilde{m}_{n,k}\exp\!\left(-\beta\lVert\widetilde{\mathbf{p}}-\widetilde{\mathbf{x}}_{n,k}\rVert_2^2\right),
    \end{aligned}
    \label{eq:motion_feature_propagation}
\end{equation}
where $\mathcal{N}_K$ indexes the $K$ trajectories with the largest raw Gaussian weights and $\beta$ is the Gaussian locality (inverse-temperature) parameter; requiring both $\widetilde{m}_{n,0}$ and $\widetilde{m}_{n,k}$ prevents tracks without a valid source or destination from contributing.
Alongside the propagated feature, we compute a presence gate from the retained trajectory mass,
\begin{equation}
    g_k(\widetilde{\mathbf{p}})
    =\operatorname{clip}_{[0,1]}\!\left(\sum_{n\in\mathcal{N}_K(\widetilde{\mathbf{p}},k)}\widetilde{w}_{n,k}(\widetilde{\mathbf{p}})\right),
    \label{eq:motion_presence_gate}
\end{equation}
which indicates where the visual condition contains trajectory-propagated appearance and allows the backbone to distinguish motion-conditioned locations from the unmodified visual context.
We denote the complete motion condition by $C_{\mathrm{motion}}=(M,g)$.
The video backbone receives $C_{\mathrm{motion}}$ as an additional visual condition while denoising the future latent sequence.
This construction follows ATI's Gaussian propagation of bilinearly sampled initial-frame features~\citep{wang_ATI_2025a} and Wan-Move's principle of replicating first-frame VAE features along latent trajectories~\citep{chu_WanMove_2025}, while our trajectories are grounded in robot commands and embodiment motion.

\begin{figure*}[t]
    \centering
    \includegraphics[width=\textwidth]{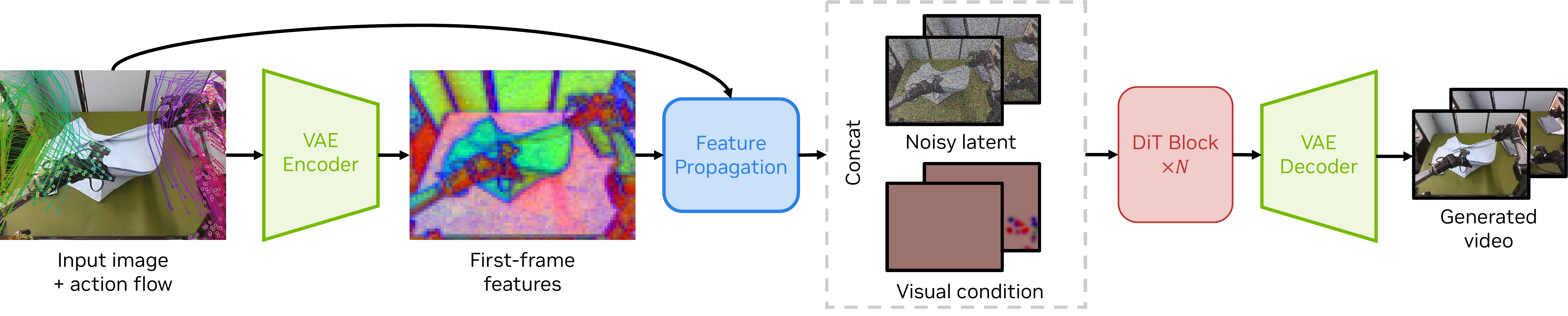}
\caption{\textbf{Action Flow as Visual Conditioning.}
Action-flow trajectories propagate first-frame latent features through time to form the motion-aware visual condition $C_{\mathrm{motion}}$.
This condition is concatenated with the noisy video latent $Z_t$ before DiT patch embedding.}
    \label{fig:motion_conditioning}
\end{figure*}

The propagated motion features are adapted to the video backbone's visual conditioning pathway while preserving the generic motion-conditioning interface, as shown in \Cref{fig:motion_conditioning}.

\paragraph{Learning objective.}
\label{sec:learning_deployment}

Let $Z_t$ denote the noised target-video latent at diffusion time $t$, let $c$ denote the pretrained text-and-image context, and let $v_t^\star$ denote the flow-matching target velocity.
We optimize the standard flow-matching objective of the video backbone while supplying the sampled motion condition:
\begin{equation}
    \mathcal{L}(\theta)=
    \mathbb{E}_{\mathbf{o}_{0:H},\mathcal{F},t,\epsilon}
    \left[\left\lVert v_\theta(Z_t,t,c,C_{\mathrm{motion}})-v^\star_t\right\rVert_2^2\right].
    \label{eq:flow_matching_objective}
\end{equation}
During training, $\mathcal{F}$ may be obtained from robot metadata or from grounding and tracking observed videos.
During deployment, $\mathcal{F}$ is instead constructed from each candidate executable command, such as a prediction from a VLA, and the model predicts that command's visual consequence.
We freeze the pretrained video-backbone weights except for the DiT patch embedding and train rank-64 low-rank adaptation (LoRA) modules~\citep{hu_lora_2021} on the $q$, $k$, $v$, and $o$ attention projections and the two feed-forward projections.
\subsection{Efficient Causal Rollout}
For efficient robotics deployment, we combine autoregressive conversion for causal long-horizon generation with improved distribution-matching distillation (DMD2)~\citep{yin_Onestep_2024,yin_improved_2024} for few-step sampling.
\paragraph{Autoregressive conversion.}
\label{sec:ar_conversion}
Joint bidirectional denoising assumes a complete action horizon and repeatedly pays full-window sampling cost, whereas long-horizon rollout benefits from reusing previously generated context.
We therefore adopt LongLive-2.0's autoregressive conversion~\citep{chen_longlive-20_2026} and partition the latent sequence into temporal chunks $\mathcal{C}_1,\ldots,\mathcal{C}_J$:
\begin{equation}
    \begin{aligned}
        &p_\theta\!\left(\mathbf{s}_{1:H}\mid\mathbf{s}_0,C_{\mathrm{motion}}\right) \\
        &\quad=\prod_{j=1}^{J}p_\theta\!\left(\mathbf{s}_{\mathcal{C}_j}\mid
        \mathbf{s}_{\mathcal{C}_{<j}},\mathbf{s}_0,C_{\mathrm{motion},\mathcal{C}_j}\right).
    \end{aligned}
    \label{eq:ar_factorization}
\end{equation}
During rollout, generated chunks replace the clean history and are reused through a KV cache.
We compute $C_{\mathrm{motion}}$ once in full-window coordinates and slice $C_{\mathrm{motion},\mathcal{C}_j}$ at absolute chunk offsets, preventing action-flow trajectories from being re-anchored at chunk boundaries.

\paragraph{Few-step distillation.}
\label{sec:few_step_distillation}
Following LongLive-2.0~\citep{chen_longlive-20_2026}, we initialize the distilled student from the causal autoregressive checkpoint and generate each chunk with four denoising steps through its causal cache while preserving the same absolute-offset motion slices.

\subsection{Open-Loop Policy Evaluation and Inverse Control}

\paragraph{Forward mode: policy evaluation.}
In forward mode, the model receives gripper flow computed from a robot command or achieved robot trajectory and predicts the resulting video outcome.
Because the target is the observed consequence rather than a prescribed behavior, any executed trajectory can provide forward-dynamics supervision regardless of task success.
Our open-loop policy evaluation uses achieved-trajectory replay: the complete recorded gripper-motion condition is fixed before generation.
More details can be found in \Cref{sec:policy_evaluation}.

\paragraph{Inverse mode: world action model.}
\label{sec:world_action_model}
We construct a world action model for the inverse mode by conditioning the model on desired object flow $\mathcal{F}^{\mathrm{obj}}$ without providing gripper flow.
The conditioning pathway also follows \Cref{eq:motion_feature_propagation}.
Desired object flow specifies task-relevant motion and conditions latent dynamics that encode compatible robot motion.
An action head trained on paired real-world robot trajectories reads clean DiT token features and learns the embodiment-specific inverse mapping from that latent representation to executable commands.

For each latent frame, the action and state decoders each concatenate mean-pooled and single-query attention-pooled spatial tokens, apply layer normalization and a two-layer GELU MLP with hidden width 1024, and predict temporally aligned robot actions and states.
The state prediction provides auxiliary supervision but is not fed back into the video model or action decoder.

Starting from the pretrained causal model, we post-train rank-32 self-attention LoRA adapters, the motion input projection, and the lightweight action and state heads on paired real-world robot trajectories.
This parameter-efficient adaptation reuses the pretrained visual-dynamics representation and restricts optimization to the rank-32 adapters, motion input projection, and lightweight readout heads.
The paired rollouts need not be expert demonstrations because each still associates achieved robot motion with the actions that produced it.
We train with
\begin{equation}
    \mathcal{L}_{\mathrm{WAM}}=\mathcal{L}_{\mathrm{flow}}+\lambda_h\left(\mathcal{L}_{\mathrm{act}}+\mathcal{L}_{\mathrm{state}}+\lambda_v\mathcal{L}_{\mathrm{vel}}\right),
    \label{eq:world_action_objective}
\end{equation}
where $\mathcal{L}_{\mathrm{act}}$ and $\mathcal{L}_{\mathrm{state}}$ are masked Huber losses on per-dimension normalized targets.
$\mathcal{L}_{\mathrm{vel}}$ is a masked L1 loss between the first differences of the predicted normalized actions and the corresponding first differences of the target normalized actions, evaluated only for consecutive pairs of valid action slots.
We use $\lambda_h=0.1$ for the combined head term and $\lambda_v=0.1$ for the inner smoothness term.
At deployment, the action and state heads predict robot actions and states from the final transformer token embeddings of each denoised video latent block, without pixel decoding.

\begin{table*}[t!]
    \centering
    \caption{\textbf{Multi-Embodiment Training Corpus.}
    Episodes and windows are counted on the annotated camera streams after resampling to 480p at 16\,fps; a window spans 81 consecutive frames ($\approx$5\,s), and windows are non-overlapping within a camera stream.
    Source-specific filtering removes static windows, frozen-gripper windows, and DROID episodes with contentless language annotations (see text).
    }
    \label{tab:datasets}
    \small
    \begin{tabular}{llrrrrl}
    \toprule
    & & & \multicolumn{2}{c}{Windows} & & \\
    \cmidrule(lr){4-5}
    Dataset & Embodiment & Episodes & before & after & Hours & License \\
    \midrule
    DROID~\cite{khazatsky_DROID_2024} & single arm & 70{,}339 & 494{,}680 & 223{,}075 & 313.7 & CC-BY-4.0 \\
    ABC-130k~\cite{allshire_scalable_2026} & bimanual arms & 54{,}961 & 1{,}060{,}456 & 1{,}048{,}681 & 1{,}474.7 & Apache-2.0 \\
    MolmoAct2~\cite{fang_molmoact2_2026} & bimanual arms & 7{,}723 & 129{,}589 & 126{,}335 & 177.7 & Apache-2.0 \\
    EgoDex~\cite{hoque_EgoDex_2025} & human hands & 41{,}888 & 89{,}380 & 89{,}380 & 125.7 & CC-BY-NC-ND-4.0 \\
    Deform360~\cite{li_deform360_2026} & handheld grippers & 1{,}714 & 87{,}224 & 60{,}315 & 84.8 & MIT \\
    XVLA-Soft-Fold~\cite{zheng_x-vla_2025} & bimanual arms & 1{,}524 & 17{,}891 & 17{,}772 & 25.0 & Apache-2.0 \\
    H1-Fold-Clothes & bimanual arms & 38 & 76 & 76 & 0.1 & Apache-2.0 \\
    \midrule
    Total & & 178{,}187 & 1{,}879{,}296 & 1{,}565{,}634 & 2{,}201.7 & \\
    \bottomrule
    \end{tabular}
\end{table*}

\section{Datasets}
\label{sec:datasets}

We assemble a multi-embodiment training corpus from seven sources that cover both robot and human interactions.
\Cref{tab:datasets} summarizes the corpus, including its size before and after our filtering and the license of each source.
DROID~\cite{khazatsky_DROID_2024} provides large-scale, scene-diverse single-arm Franka manipulation and is the primary robot source.
ABC-130k~\cite{allshire_scalable_2026} and MolmoAct2~\cite{fang_molmoact2_2026} provide bimanual teleoperation.
EgoDex~\cite{hoque_EgoDex_2025} provides egocentric human-hand manipulation captured with Apple Vision Pro, while Deform360~\cite{li_deform360_2026} provides handheld-gripper demonstrations of deformable-object manipulation using UMI grippers~\cite{chi_Universal_2024}.
XVLA-Soft-Fold~\cite{zheng_x-vla_2025} and H1-Fold-Clothes\footnote{\url{https://huggingface.co/datasets/lerobot/unitreeh1_fold_clothes}} contribute smaller bimanual and humanoid cloth-folding sets.
The Interactive World Simulator (IWS) tasks~\cite{wang_Interactive_2026} are used only for the data-efficiency evaluations and are excluded from the training corpus.

\paragraph{Preprocessing and annotation.}
Because interaction with deformable objects is the focus of this paper, we keep mainly deformable-object interaction data (cloth, cable, rope, bag, and paper).
The main corpus sources are converted to a shared per-episode, per-camera layout: frames are resampled to 480p at 16\,fps and cut into non-overlapping 81-frame windows, with dense point tracks, a language caption, precomputed VAE latents, and source-dependent embodiment and object annotations where available.
Dense tracks come from AllTracker~\cite{harley_AllTracker_2025} on a $128\times128$ query grid per window.
Depending on the source, embodiment masks come from calibrated robot rendering, projected hand geometry, SAM~3 refinement, or text prompting~\cite{carion_SAM_2025}.
Object masks use either source-provided refined masks or text-prompted SAM~3 and are unavailable for DROID and EgoDex.
Captions are generated per episode using dataset-provided captions or a vision-language model.

\paragraph{Filtering.}
Beyond the task-level deformable subsetting above, we apply source-specific combinations of three window-level filters, with thresholds tuned from motion-score histograms.
A \emph{static-window} filter drops windows whose 90th-percentile visible track path length is below 50 pixels at 480p.
A \emph{frozen-gripper} filter drops windows whose embodiment tracks fall below the same 50-pixel floor.
A \emph{contentless-caption} filter drops DROID episodes whose language annotation describes no actionable content.
For wrist-view coverage, we add a separate DROID wrist-camera stream of 17-frame windows restricted to deformable-object episodes identified from language annotations; it is used as a standalone viewpoint and is not combined with the external-camera views.

\begin{figure*}[t!]
    \centering
    \includegraphics[width=\textwidth]{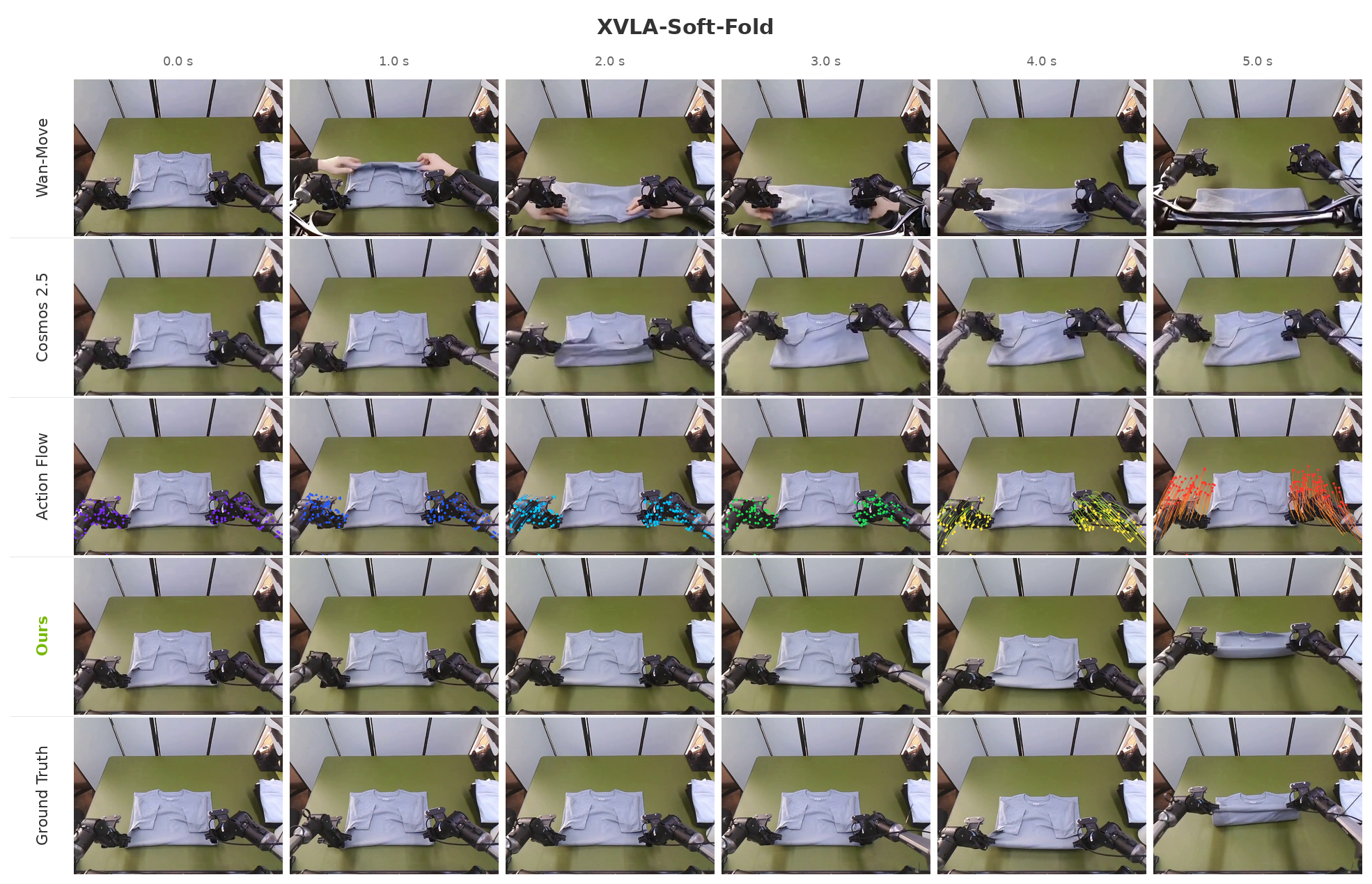}
    \par\vspace{0.8em}
    \includegraphics[width=\textwidth]{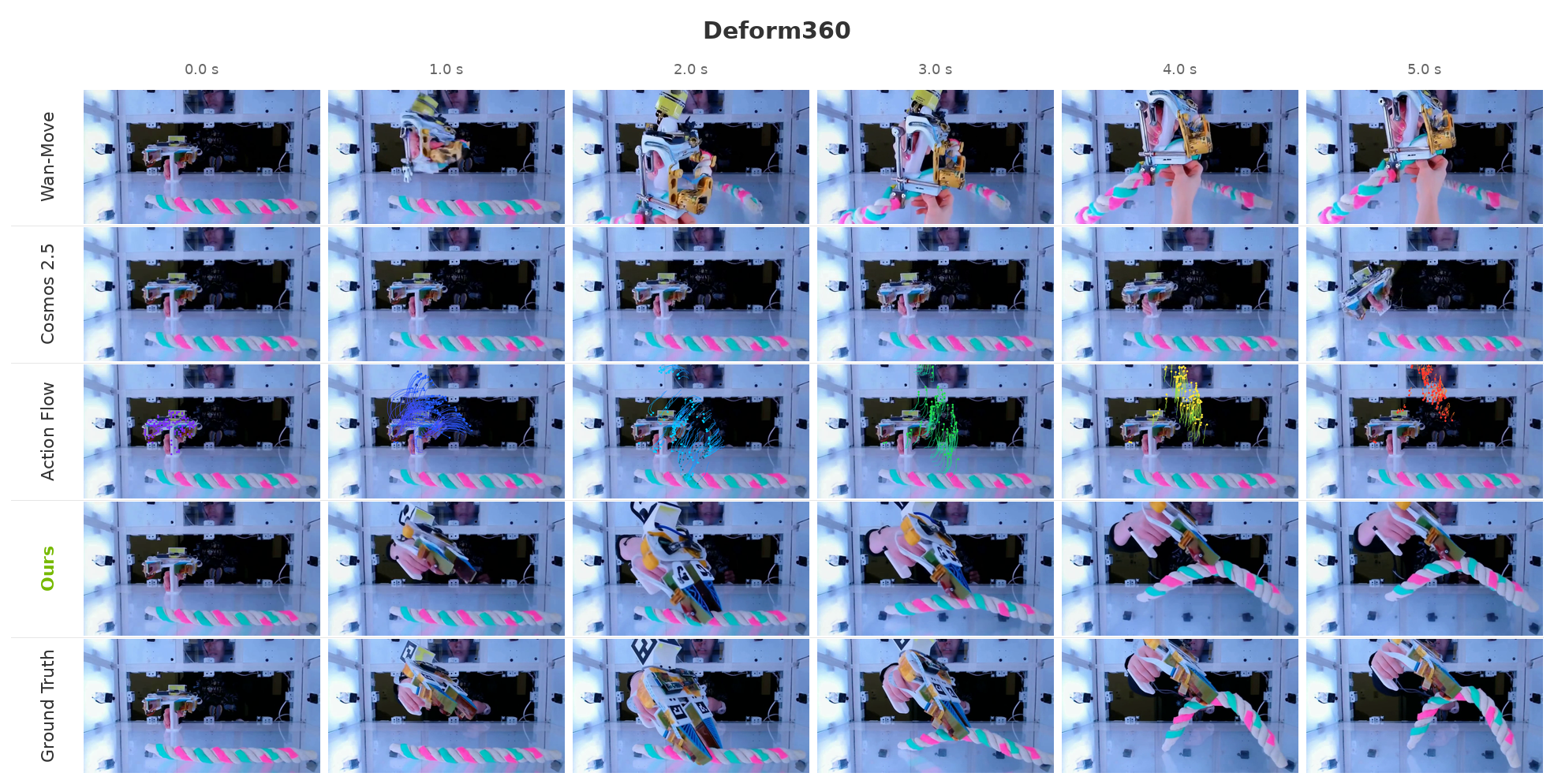}
    \caption{\textbf{Qualitative Evaluation on XVLA-Soft-Fold and Deform360.}
    Each dataset-specific panel shows six synchronized snapshots over the full five-second horizon.
    From top to bottom, the rows show \texttt{Wan-Move}, fine-tuned \texttt{Cosmos 2.5}, the action-flow conditioning input, our clean generated output, and the matched ground truth.
    Additional results appear in the appendix.
    }
    \label{fig:qualitative_checkpoint_eval}
\end{figure*}

\section{Experiments}

We organize the experiments to test whether action flow supports transferable prediction, efficient rollout, open-loop policy evaluation, and inverse control.
The primary target domain is robot interaction with deformable objects, for which robust explicit state reconstruction and physics simulation remain challenging.
We adopt \texttt{Cosmos 2.5} as our baseline, which uses its native relative 6D end-effector action representation.
For comparison, we equip both Cosmos 2.5 and Wan2.2 with the same image-space action-flow representation and train them on the same robot-motion trajectories and dataset mixture.

\subsection{Implementation Details}
\label{sec:implementation_details}

We implement action flow using different video backbones: Cosmos 2.5~\cite{nvidia_World_2026} and two variants of Wan2.2~\cite{wan_Wan_2025}, I2V-A14B and TI2V-5B.
All three backbones use the same track sampling and motion-feature aggregation described below; only the adapter that injects the resulting motion condition into the video model differs.

\paragraph{Cosmos 2.5.}
Because the Cosmos 2.5 latent has 16 channels, we form a 17-channel motion side input from 16 propagated first-frame feature channels and one presence-gate channel.
A separate zero-initialized patch projection embeds this side input, and the resulting tokens are added to the noisy-video token embeddings at every denoising call.

\paragraph{Wan2.2 I2V-A14B.}
The visual condition contains 16 VAE-feature channels and four mask channels: latent frame zero retains the native I2V condition, later frames receive the raw propagated feature mass plus the pristine feature scaled by one minus the presence gate, and the gate is broadcast across the four mask channels.
The resulting visual condition is concatenated with the noisy latent at the DiT input.
Following Wan2.2's two-expert MoE schedule, the motion-conditioned features are supplied to the motion-fine-tuned high-noise expert during early denoising; at the MoE transition, the pristine visual condition is restored and denoising continues with the unmodified low-noise expert.

\paragraph{Wan2.2 TI2V-5B.}
Because TI2V-5B does not use the I2V visual-conditioning tensor, we concatenate a 49-channel motion side input (48 propagated feature channels and one presence-gate channel) with the noisy latent at every denoising call and widen the DiT input projection accordingly.
This side input remains constant across denoising steps and is neither noised nor used to re-anchor the latent state.

\begin{table*}[t!]
\centering
\caption{\textbf{Evaluation on Multi-Embodiment Datasets.}
All methods use the same 100 validation clips per dataset across five held-out sets.
PSNR and SSIM measure full-reference fidelity; object and gripper EPE measure tracked-point motion error in pixels; FID and FVD measure frame and video distribution quality; VLM is scored on a 1--5 scale.
Average is the unweighted mean across available dataset strata.
Best results within each dataset and the Average block are bold.
Gray rows denote zero-shot baselines.
DROID object EPE is omitted because reliable manipulated-object masks cannot be obtained in its cluttered scenes.}
\label{tab:checkpoint_eval}
\scriptsize
\setlength{\tabcolsep}{3.2pt}
\begin{adjustbox}{max width=\textwidth}
\begin{tabular}{llrrrrrrr}
\toprule
Dataset & Model & PSNR$\uparrow$ & SSIM$\uparrow$ & Obj.\ EPE$\downarrow$ & Grip.\ EPE$\downarrow$ & FID$\downarrow$ & FVD$\downarrow$ & VLM$\uparrow$ \\
\midrule
\rowcolor{gray!15} \cellcolor{white}\multirow{7}{*}{XVLA-Soft-Fold} & ATI & $14.47_{\pm 0.29}$ & $0.653_{\pm 0.009}$ & $15.84_{\pm 3.24}$ & $4.61_{\pm 0.25}$ & $29.8$ & $533.6$ & $3.58_{\pm 0.13}$ \\
\rowcolor{gray!15} \cellcolor{white} & Wan-Move & $12.82_{\pm 0.25}$ & $0.614_{\pm 0.010}$ & $24.06_{\pm 3.91}$ & $4.88_{\pm 0.35}$ & $31.7$ & $681.1$ & $3.65_{\pm 0.13}$ \\
 & Cosmos 2.5 & $14.95_{\pm 0.21}$ & $0.677_{\pm 0.006}$ & $6.65_{\pm 1.46}$ & $35.78_{\pm 2.01}$ & $36.0$ & $479.9$ & $4.56_{\pm 0.05}$ \\
 & Ours (Cosmos 2.5 2B) & $16.62_{\pm 0.28}$ & $0.719_{\pm 0.007}$ & $5.29_{\pm 1.65}$ & $17.36_{\pm 1.93}$ & $32.1$ & $378.5$ & $4.42_{\pm 0.08}$ \\
 & Ours (Wan2.2 5B) & $17.89_{\pm 0.28}$ & $0.738_{\pm 0.007}$ & $\mathbf{3.40}_{\pm 0.67}$ & $3.80_{\pm 0.20}$ & $21.3$ & $345.1$ & $4.53_{\pm 0.05}$ \\
 & Ours (Wan2.2 A14B) & $18.25_{\pm 0.26}$ & $0.746_{\pm 0.006}$ & $3.42_{\pm 0.65}$ & $4.17_{\pm 0.25}$ & $19.8$ & $255.0$ & $4.58_{\pm 0.06}$ \\
 & Ours (Wan2.2 A14B 4-step) & $\mathbf{19.23}_{\pm 0.29}$ & $\mathbf{0.781}_{\pm 0.006}$ & $3.47_{\pm 0.64}$ & $\mathbf{3.20}_{\pm 0.16}$ & $\mathbf{19.3}$ & $\mathbf{238.5}$ & $\mathbf{4.67}_{\pm 0.05}$ \\
\midrule
\rowcolor{gray!15} \cellcolor{white}\multirow{7}{*}{Deform360} & ATI & $15.56_{\pm 0.34}$ & $0.643_{\pm 0.014}$ & $30.15_{\pm 7.55}$ & $6.85_{\pm 0.56}$ & $48.9$ & $569.1$ & $2.66_{\pm 0.15}$ \\
\rowcolor{gray!15} \cellcolor{white} & Wan-Move & $16.11_{\pm 0.30}$ & $0.676_{\pm 0.012}$ & $21.13_{\pm 2.80}$ & $7.05_{\pm 0.64}$ & $42.9$ & $389.1$ & $3.54_{\pm 0.14}$ \\
 & Cosmos 2.5 & $15.86_{\pm 0.26}$ & $0.630_{\pm 0.009}$ & $20.02_{\pm 2.46}$ & $47.85_{\pm 3.20}$ & $51.1$ & $474.5$ & $2.90_{\pm 0.14}$ \\
 & Ours (Cosmos 2.5 2B) & $17.53_{\pm 0.30}$ & $0.671_{\pm 0.008}$ & $8.47_{\pm 1.83}$ & $21.53_{\pm 3.28}$ & $41.6$ & $295.0$ & $2.67_{\pm 0.14}$ \\
 & Ours (Wan2.2 5B) & $18.53_{\pm 0.28}$ & $0.751_{\pm 0.007}$ & $9.29_{\pm 0.86}$ & $6.02_{\pm 0.47}$ & $30.3$ & $228.3$ & $3.06_{\pm 0.14}$ \\
 & Ours (Wan2.2 A14B) & $19.50_{\pm 0.30}$ & $0.786_{\pm 0.007}$ & $7.69_{\pm 1.02}$ & $5.68_{\pm 0.50}$ & $29.1$ & $213.0$ & $2.90_{\pm 0.15}$ \\
 & Ours (Wan2.2 A14B 4-step) & $\mathbf{20.63}_{\pm 0.32}$ & $\mathbf{0.804}_{\pm 0.007}$ & $\mathbf{6.86}_{\pm 0.81}$ & $\mathbf{5.06}_{\pm 0.42}$ & $\mathbf{25.5}$ & $\mathbf{166.5}$ & $\mathbf{3.70}_{\pm 0.13}$ \\
\midrule
\rowcolor{gray!15} \cellcolor{white}\multirow{7}{*}{DROID} & ATI & $17.27_{\pm 0.39}$ & $0.706_{\pm 0.016}$ & \textemdash & $4.00_{\pm 0.27}$ & $40.4$ & $344.9$ & $3.08_{\pm 0.15}$ \\
\rowcolor{gray!15} \cellcolor{white} & Wan-Move & $16.03_{\pm 0.33}$ & $0.658_{\pm 0.015}$ & \textemdash & $4.24_{\pm 0.32}$ & $41.2$ & $349.5$ & $3.65_{\pm 0.14}$ \\
 & Cosmos 2.5 & $16.21_{\pm 0.21}$ & $0.651_{\pm 0.007}$ & \textemdash & $28.45_{\pm 2.60}$ & $39.9$ & $295.4$ & $4.04_{\pm 0.11}$ \\
 & Ours (Cosmos 2.5 2B) & $19.50_{\pm 0.21}$ & $0.717_{\pm 0.006}$ & \textemdash & $3.97_{\pm 0.25}$ & $30.0$ & $179.7$ & $4.09_{\pm 0.11}$ \\
 & Ours (Wan2.2 5B) & $19.22_{\pm 0.26}$ & $0.753_{\pm 0.009}$ & \textemdash & $3.55_{\pm 0.20}$ & $28.8$ & $228.9$ & $3.85_{\pm 0.14}$ \\
 & Ours (Wan2.2 A14B) & $21.14_{\pm 0.24}$ & $0.826_{\pm 0.005}$ & \textemdash & $3.55_{\pm 0.25}$ & $21.7$ & $163.2$ & $4.14_{\pm 0.11}$ \\
 & Ours (Wan2.2 A14B 4-step) & $\mathbf{22.10}_{\pm 0.26}$ & $\mathbf{0.851}_{\pm 0.005}$ & \textemdash & $\mathbf{3.13}_{\pm 0.20}$ & $\mathbf{19.0}$ & $\mathbf{118.6}$ & $\mathbf{4.26}_{\pm 0.10}$ \\
\midrule
\rowcolor{gray!15} \cellcolor{white}\multirow{7}{*}{MolmoAct2} & ATI & $17.18_{\pm 0.33}$ & $0.706_{\pm 0.011}$ & $29.75_{\pm 6.64}$ & $4.02_{\pm 0.20}$ & $28.4$ & $414.0$ & $3.16_{\pm 0.12}$ \\
\rowcolor{gray!15} \cellcolor{white} & Wan-Move & $17.67_{\pm 0.28}$ & $0.734_{\pm 0.009}$ & $13.85_{\pm 2.43}$ & $3.91_{\pm 0.24}$ & $21.6$ & $293.5$ & $3.68_{\pm 0.13}$ \\
 & Cosmos 2.5 & $14.03_{\pm 0.15}$ & $0.618_{\pm 0.006}$ & $12.24_{\pm 1.47}$ & $33.30_{\pm 1.65}$ & $29.8$ & $476.3$ & $3.71_{\pm 0.11}$ \\
 & Ours (Cosmos 2.5 2B) & $17.24_{\pm 0.31}$ & $0.698_{\pm 0.007}$ & $5.36_{\pm 0.85}$ & $15.47_{\pm 1.56}$ & $25.5$ & $307.5$ & $3.73_{\pm 0.10}$ \\
 & Ours (Wan2.2 5B) & $20.16_{\pm 0.22}$ & $0.776_{\pm 0.005}$ & $7.06_{\pm 1.01}$ & $3.21_{\pm 0.14}$ & $18.7$ & $240.2$ & $3.81_{\pm 0.09}$ \\
 & Ours (Wan2.2 A14B) & $21.50_{\pm 0.23}$ & $0.806_{\pm 0.005}$ & $6.43_{\pm 0.77}$ & $2.97_{\pm 0.15}$ & $14.3$ & $203.0$ & $3.94_{\pm 0.11}$ \\
 & Ours (Wan2.2 A14B 4-step) & $\mathbf{22.63}_{\pm 0.24}$ & $\mathbf{0.837}_{\pm 0.005}$ & $\mathbf{5.22}_{\pm 0.64}$ & $\mathbf{2.63}_{\pm 0.14}$ & $\mathbf{12.3}$ & $\mathbf{147.5}$ & $\mathbf{4.14}_{\pm 0.10}$ \\
\midrule
\rowcolor{gray!15} \cellcolor{white}\multirow{7}{*}{ABC-130k} & ATI & $20.55_{\pm 0.28}$ & $0.790_{\pm 0.008}$ & $17.01_{\pm 2.95}$ & $3.63_{\pm 0.17}$ & $34.6$ & $359.3$ & $3.19_{\pm 0.14}$ \\
\rowcolor{gray!15} \cellcolor{white} & Wan-Move & $19.10_{\pm 0.26}$ & $0.758_{\pm 0.010}$ & $27.08_{\pm 9.62}$ & $3.29_{\pm 0.16}$ & $34.7$ & $328.1$ & $4.04_{\pm 0.11}$ \\
 & Cosmos 2.5 & $17.07_{\pm 0.16}$ & $0.764_{\pm 0.005}$ & $14.00_{\pm 2.20}$ & $26.05_{\pm 1.53}$ & $38.9$ & $303.0$ & $4.16_{\pm 0.10}$ \\
 & Ours (Cosmos 2.5 2B) & $21.14_{\pm 0.37}$ & $0.823_{\pm 0.006}$ & $5.98_{\pm 0.68}$ & $10.65_{\pm 1.11}$ & $32.6$ & $226.4$ & $4.23_{\pm 0.10}$ \\
 & Ours (Wan2.2 5B) & $22.40_{\pm 0.20}$ & $0.833_{\pm 0.004}$ & $6.70_{\pm 0.77}$ & $2.82_{\pm 0.12}$ & $21.6$ & $201.5$ & $4.26_{\pm 0.09}$ \\
 & Ours (Wan2.2 A14B) & $23.43_{\pm 0.23}$ & $0.859_{\pm 0.004}$ & $6.45_{\pm 0.79}$ & $2.80_{\pm 0.12}$ & $18.9$ & $134.2$ & $4.36_{\pm 0.08}$ \\
 & Ours (Wan2.2 A14B 4-step) & $\mathbf{24.62}_{\pm 0.23}$ & $\mathbf{0.880}_{\pm 0.004}$ & $\mathbf{5.53}_{\pm 0.71}$ & $\mathbf{2.45}_{\pm 0.11}$ & $\mathbf{17.5}$ & $\mathbf{108.6}$ & $\mathbf{4.37}_{\pm 0.08}$ \\
\midrule
\rowcolor{gray!15} \cellcolor{white}\multirow{7}{*}{Average} & ATI & $17.01_{\pm 0.15}$ & $0.700_{\pm 0.005}$ & $23.19_{\pm 2.74}$ & $4.62_{\pm 0.14}$ & $36.4$ & $444.2$ & $3.14_{\pm 0.06}$ \\
\rowcolor{gray!15} \cellcolor{white} & Wan-Move & $16.35_{\pm 0.13}$ & $0.688_{\pm 0.005}$ & $21.53_{\pm 2.76}$ & $4.67_{\pm 0.17}$ & $34.4$ & $408.3$ & $3.71_{\pm 0.06}$ \\
 & Cosmos 2.5 & $15.62_{\pm 0.09}$ & $0.668_{\pm 0.003}$ & $13.23_{\pm 0.97}$ & $34.28_{\pm 1.02}$ & $39.1$ & $405.8$ & $3.88_{\pm 0.05}$ \\
 & Ours (Cosmos 2.5 2B) & $18.41_{\pm 0.13}$ & $0.725_{\pm 0.003}$ & $6.27_{\pm 0.67}$ & $13.80_{\pm 0.85}$ & $32.4$ & $277.4$ & $3.83_{\pm 0.05}$ \\
 & Ours (Wan2.2 5B) & $19.64_{\pm 0.11}$ & $0.770_{\pm 0.003}$ & $6.61_{\pm 0.42}$ & $3.88_{\pm 0.12}$ & $24.1$ & $248.8$ & $3.90_{\pm 0.05}$ \\
 & Ours (Wan2.2 A14B) & $20.76_{\pm 0.11}$ & $0.805_{\pm 0.003}$ & $6.00_{\pm 0.41}$ & $3.83_{\pm 0.13}$ & $20.7$ & $193.7$ & $3.98_{\pm 0.05}$ \\
 & Ours (Wan2.2 A14B 4-step) & $\mathbf{21.84}_{\pm 0.12}$ & $\mathbf{0.830}_{\pm 0.002}$ & $\mathbf{5.27}_{\pm 0.35}$ & $\mathbf{3.29}_{\pm 0.10}$ & $\mathbf{18.7}$ & $\mathbf{155.9}$ & $\mathbf{4.23}_{\pm 0.04}$ \\
\bottomrule
\end{tabular}
\end{adjustbox}
\end{table*}

\paragraph{Training.}
We sample one of four conditioning modes (\emph{None}, \emph{Embodiment}, \emph{Object}, or \emph{All}) with canonical I2V probabilities $(0.05,0.40,0.40,0.15)$, respectively; when a required pool is unavailable, we remove that mode and renormalize the remaining probabilities, while dataset- and camera-specific overrides accommodate unreliable or unavailable annotations and viewpoints.
In Embodiment or Object mode, we sample $1$--$128$ tracks from the corresponding pool; in All mode, we sample $256$--$1024$ tracks from the union of embodiment, object, and unassigned scene tracks.

AllTracker trajectories are pooled into latent-frame windows, with visibility defined by any visible sample in the window and destination position defined as the mean visible position.
We use normalized latent-grid coordinates, Gaussian locality $\beta=220$, and the two largest raw trajectory weights at each destination cell; the propagated feature is their unnormalized weighted sum, and their raw-weight sum is clipped to $[0,1]$ to form the presence gate.

We build on pretrained Wan2.2 and Cosmos 2.5 video backbones, which provide general video-generation priors.
Starting from these checkpoints, we perform multi-embodiment action-flow mid-training to learn a shared action-conditioning interface across embodiments.
We train our Wan2.2 I2V-A14B backbone for five days and 40,000 steps using 32 NVIDIA H100 GPUs; the other backbones require fewer computational resources.

\paragraph{Metrics.} Following prior evaluations of action-conditioned world models and trajectory-controlled video generation~\citep{wang_Interactive_2026,guo_CtrlWorld_2025,wang_ATI_2025a,chu_WanMove_2025}, we report PSNR and SSIM as full-reference fidelity metrics, LPIPS~\cite{zhang_unreasonable_2018} as perceptual discrepancy, object and gripper endpoint error (EPE) as motion error, and FID~\cite{heusel_gans_2017} and FVD~\cite{unterthiner_towards_2019} as frame- and video-distribution discrepancy.
We additionally report a deterministic VLM-as-a-judge score using the Gemma~4 31B instruction-tuned model~\citep{team_gemma_2026}; implementation details are provided in \Cref{app:vlm_judge}.

\subsection{Effectiveness of Actionable Flow Conditioning}
We test whether actionable flow transfers across video-generation architectures by training action flow-conditioned Cosmos 2.5 and Wan2.2 models with the same flow construction, dataset mixture, and prediction horizon.
The evaluated backbones are Cosmos 2.5 and the Wan2.2 I2V-A14B and TI2V-5B variants, including a four-denoising-step LightX2V I2V-A14B checkpoint~\citep{noauthor_lightx2v_2026}.
We evaluate all methods on the same held-out clips from each of five validation sets: XVLA-Soft-Fold, Deform360, DROID, MolmoAct2, and ABC-130k.
We fine-tune the \texttt{Cosmos 2.5} baseline using its native relative 6D end-effector action representation.
\texttt{ATI} and \texttt{Wan-Move} are evaluated using their released checkpoints without multi-embodiment mid-training and are shown as gray rows in \Cref{tab:checkpoint_eval}.

\paragraph{Results.}
\texttt{Ours (Wan2.2 A14B 4-step)} has the best displayed point estimate in most cells of \Cref{tab:checkpoint_eval}.
Within the same Cosmos 2.5 backbone, replacing the native relative 6D action in \texttt{Cosmos 2.5} with actionable flow in \texttt{Ours (Cosmos 2.5 2B)} yields better displayed point estimates for PSNR, SSIM, gripper-flow EPE, FID, and FVD on all five datasets, as well as object-flow EPE on all four datasets where object EPE is reported.
This controlled comparison isolates the conditioning representation and shows that image-space actionable flow communicates embodied motion more effectively than the native relative 6D action.
VLM scores are mixed because they assess physical plausibility, temporal consistency, object permanence, and motion realism rather than whether the generated motion precisely follows the commanded gripper and object trajectories, which are measured directly by flow EPE.

\begin{figure}[t]
    \centering
    \includegraphics[width=\columnwidth]{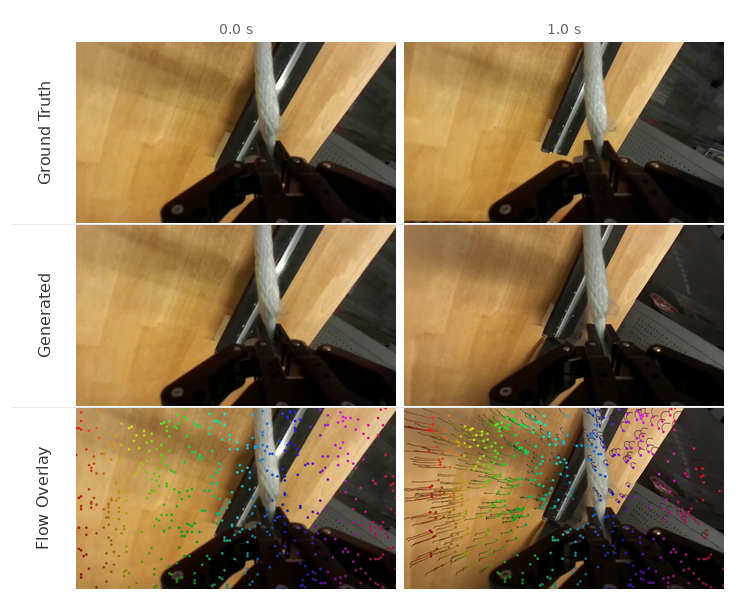}
    \caption{\textbf{Wrist-Camera Egomotion Prediction.}
    The first and last synchronized snapshots show frames 0 and 16 of the 17-frame generated clip, corresponding to 0.0 and 1.0 seconds of physical time.
    }
    \label{fig:droid_wrist_camera_egomotion}
\end{figure}

ATI and Wan-Move achieve nontrivial zero-shot performance, indicating that generic trajectory-conditioned video models can transfer to unseen robot tasks.
However, their object-flow EPE and FVD point estimates are higher than those of our best model, leaving a measurable performance gap.
Together, the Cosmos 2.5 and Wan2.2 results support transfer of actionable-flow conditioning across the evaluated backbone families.
Qualitatively, \Cref{fig:qualitative_checkpoint_eval} compares matched XVLA-Soft-Fold and Deform360 validation rollouts, with the additional DROID comparison in \Cref{fig:qualitative_checkpoint_eval_droid}.
We additionally test whether the same flow representation can accommodate wrist-camera observations, where an arm-mounted camera produces global image motion as the robot moves.
Given depth and relative camera pose, camera egomotion can be projected into image-space flow and combined with interaction motion; \Cref{fig:droid_wrist_camera_egomotion} provides a qualitative proof of concept on DROID.

\begin{figure*}[t]
    \centering
    \includegraphics[width=\textwidth]{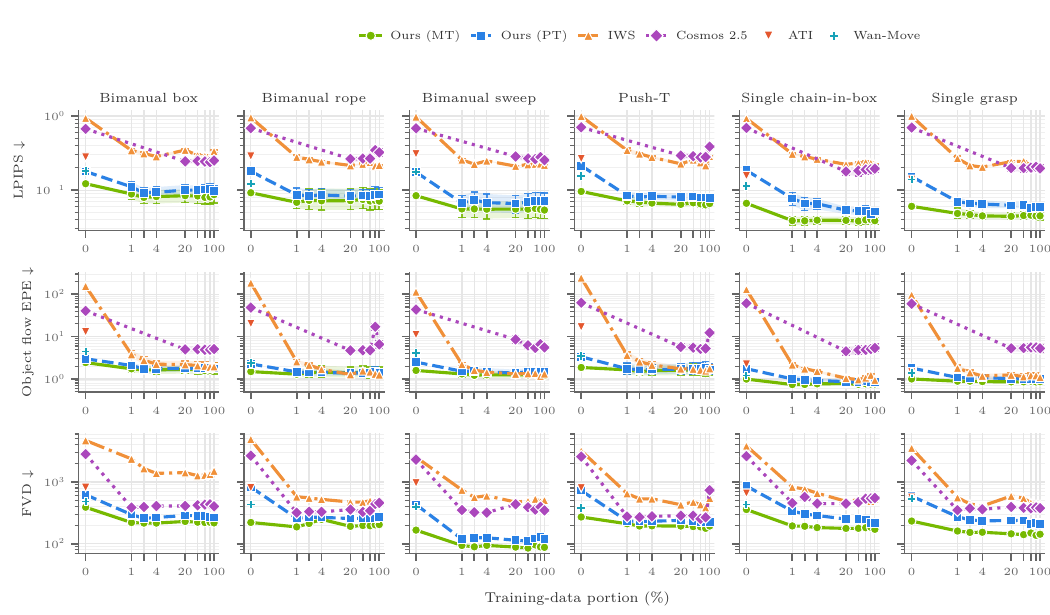}
    \caption{\textbf{Data Efficiency on IWS Tasks.}
    Each task reports LPIPS, object-flow endpoint error (EPE), and FVD versus the percentage of task-specific training data.
    For all metrics, lower is better.
    \texttt{Ours (PT)} starts from the original Wan2.2 weights, whereas \texttt{Ours (MT)} starts from our multi-embodiment mid-trained checkpoint.
    \texttt{Wan-Move} and \texttt{ATI} are evaluated zero-shot.}
    \label{fig:iws_data_efficiency}
\end{figure*}

\subsection{Data Efficiency from Multi-Embodiment Mid-Training}
Our action flow enables efficient multi-embodiment mid-training.
We test whether our multi-embodiment mid-training reduces the amount of task-specific robot data needed for post-training.
We use six tasks from the Interactive World Simulator (IWS) dataset~\cite{wang_Interactive_2026}: Bimanual box, Bimanual rope, Bimanual sweep, Push-T, Single chain-in-box, and Single grasp.
The IWS dataset is excluded from multi-embodiment mid-training; during task-specific adaptation, each method uses the same nested subset of the common eligible training cohort.
For each task, we form seeded, nested 1\%, 2\%, 4\%, 20\%, 40\%, 60\%, 80\%, and 100\% subsets of that cohort and use the same subset for every method; 0\% denotes zero-shot evaluation with no target-task adaptation.
We denote the model initialized directly from pretrained Wan2.2 weights as \texttt{Ours (PT)} and the model initialized after our multi-embodiment action-flow mid-training as \texttt{Ours (MT)}.
We compare both variants with \texttt{IWS}, which trains a task-specific action-conditioned world model from scratch, and with the original \texttt{Cosmos 2.5} baseline under the same data fractions.
At the 0\% data point, we additionally compare with \texttt{Wan-Move} and \texttt{ATI} to test whether generic motion-conditioned video models transfer zero-shot to the same IWS tasks.

\paragraph{Results.}
As shown in \Cref{fig:iws_data_efficiency}, our multi-embodiment mid-training provides the strongest transfer before seeing any task-specific data.
At 0\%, \texttt{Ours (MT)} has better LPIPS, object-flow EPE, and FVD point estimates than \texttt{Ours (PT)} across all six tasks, a pattern consistent with transfer from multi-embodiment mid-training.
The task-specific \texttt{IWS} model and the newly introduced action-conditioning layers of \texttt{Cosmos 2.5} are substantially worse on all three metrics at 0\%, indicating that their unadapted action interfaces do not transfer effectively to these held-out tasks.
The generic motion-conditioned baselines can produce plausible motion, but their weaker object-flow accuracy relative to \texttt{Ours (MT)} shows that generic motion control alone does not ensure faithful execution of commanded object trajectories.

The advantage persists after adaptation.
At 100\%, \texttt{Ours (MT)} has the lowest LPIPS and FVD on all six tasks and the lowest flow EPE on four; on Bimanual box it is within 0.014 pixels of \texttt{Ours (PT)}, and on Bimanual rope it is within 0.15 pixels of \texttt{IWS}.
\texttt{Ours (PT)} is consistently second-best in FVD, showing that the original video prior is useful, while the stronger results of \texttt{Ours (MT)} are consistent with transferable interaction dynamics beyond appearance quality.
Most of \texttt{Ours (MT)}'s task-specific gains arrive by the 20\% fraction: between 20\% and 100\%, its per-task values change by at most 3.4\% in LPIPS, 6.7\% in flow EPE, and 6.8\% in FVD.
Because FVD is estimated from 40 clips per point and has high variance, its small, occasionally non-monotonic changes beyond 20\% do not establish strict convergence.
Nevertheless, the early performance plateau across complementary appearance, motion, and temporal metrics suggests that multi-embodiment mid-training can reduce the target-task data needed for action-flow post-training.

\subsection{Inference Speed}
\label{sec:inference_speed}

Robotics deployment motivates measuring generation latency and throughput to assess whether rollouts fit within an interaction budget.

\paragraph{Protocol.}
We benchmark the stages of \Cref{sec:ar_conversion,sec:few_step_distillation} on one H100 GPU in bfloat16 at batch size one, generating 81 frames at $480\times832$ (21 latent frames, corresponding to a $30\times52$ DiT patch-token grid).
With guidance and VAE decoding disabled, we report generation-only throughput as the median of five clips after three warm-ups.
These measurements use a short, three-chunk horizon where the 24-latent attention cap is inactive and long-rollout costs are not exercised.
The separately run bidirectional baseline also uses a quantized attention kernel.

\paragraph{Results.}
Relative to full-sequence bidirectional sampling, cached causal autoregressive conversion is $1.68\times$ faster despite increasing network evaluations from 50 to 153 because each processes a seven-latent chunk, including one recache per chunk (\Cref{tab:inference_speed}).
Distillation reduces evaluations to 15 for a further $\approx9.5\times$ gain, reaching 62.0 frames per second and a $16.0\times$ generation-only speedup.
This generation-only value excludes guidance and VAE/pixel decoding and is not an end-to-end speedup.
Separately, the world-action readout of \Cref{sec:world_action_model} consumes internal DiT features without decoding, but the reported world-action model uses the 50-step autoregressive teacher rather than the few-step student.

\begin{table}[t!]
\centering
\caption{\textbf{Inference Speed Across Optimization Stages.}
Measurements use a single 80-GB H100 GPU with bfloat16 and a batch size of one to generate 81 frames at $480\times832$ from a single conditioning frame.
Guidance is disabled, and VAE decoding time is excluded.
We report the median over five timed clips following three warm-up clips.
Both teacher models use 50 sampling steps, whereas the student uses four.
Speedup is calculated relative to the bidirectional teacher.}
\label{tab:inference_speed}
\small
\setlength{\tabcolsep}{5pt}
\begin{tabular}{lrrr}
\toprule
Stage & s\,/\,clip$\downarrow$ & FPS$\uparrow$ & Speedup$\uparrow$ \\
\midrule
Bidirectional teacher & 20.92 & 3.87 & 1.0$\times$ \\
Autoregressive teacher & 12.48 & 6.49 & 1.68$\times$ \\
Few-step student & $\mathbf{1.31}$ & $\mathbf{61.98}$ & $\mathbf{16.0\times}$ \\
\bottomrule
\end{tabular}
\end{table}

\subsection{Open-Loop Policy Evaluation}
\label{sec:policy_evaluation}

\paragraph{Protocol.}
We evaluate whether Hydra-0 preserves policy outcomes under open-loop replay in RoboLab~\cite{yang_RoboLab_2026}.
Starting from each rollout's true first observation, Hydra-0 receives causal action flow derived from the recorded robot end-effector trajectory; the policy is not queried on generated observations for open-loop evaluation.
The RoboLab study evaluates five pre-trained policies on six tasks with 10 rollouts per policy--task pair (300 episodes), comparing environment success with judgments from a human rater blinded to policy identity.
Following prior neural-simulator evaluations~\cite{wang_Interactive_2026}, we compare reference and generated success rates across policy--task pairs using Pearson correlation, Spearman correlation, and mean absolute error.

\begin{figure}[t!]
    \centering
    \includegraphics[width=\columnwidth]{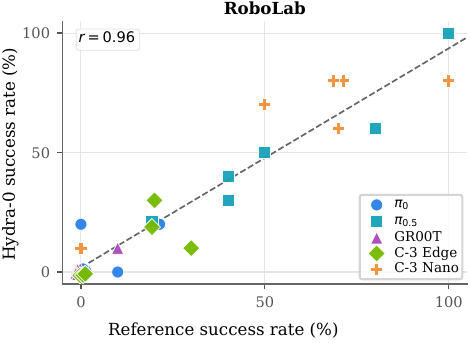}
    \caption{\textbf{Open-Loop Policy Evaluation with \shortname.}
    Each point aggregates 10 trials; dashed lines show least-squares fits.
    We evaluate $\pi_0$~\cite{black_$pi_0$_2024}, $\pi_{0.5}$~\cite{intelligence_$p_05$_2025}, GR00T N1.7~\cite{nvidia_GR00T_2025}, Cosmos-3 Edge, and Cosmos-3 Nano~\cite{nvidia_cosmos_2026} in RoboLab.
    Legend abbreviations: GR00T denotes GR00T N1.7; C-3 Edge and C-3 Nano denote Cosmos-3 Edge and Cosmos-3 Nano.
    }
    \label{fig:policy_evaluation_correlation}
\end{figure}

\begin{figure}[t!]
    \centering
    \includegraphics[width=\columnwidth]{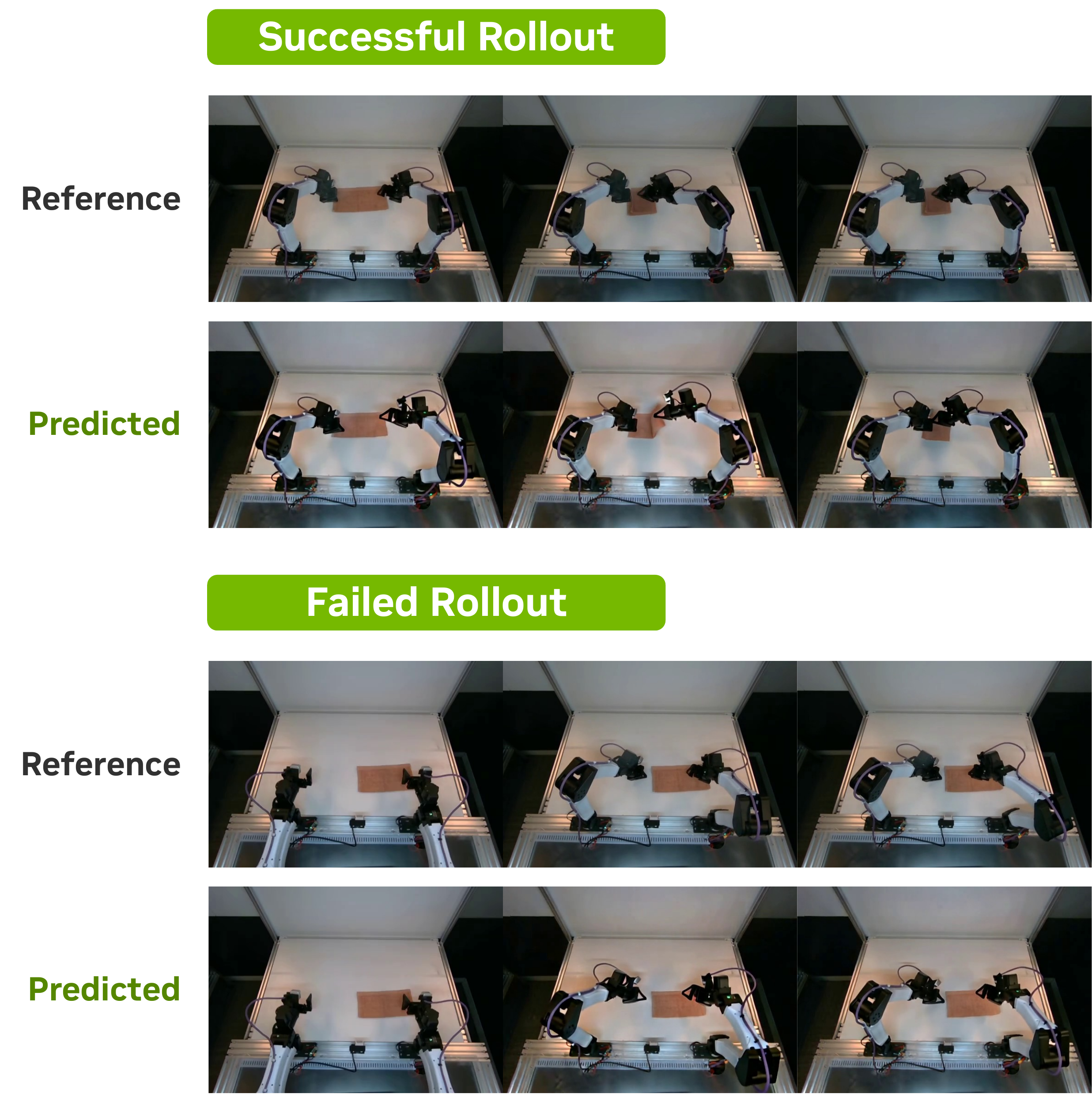}
    \caption{\textbf{Real-World Open-Loop Policy Evaluation.}
    Qualitative cloth-folding rollouts show the recorded reference trajectory and the corresponding open-loop prediction conditioned on the recorded robot motion.
    }
    \label{fig:real_world_policy_evaluation}
\end{figure}

\begin{figure*}[t!]
    \centering
    \includegraphics[width=\textwidth]{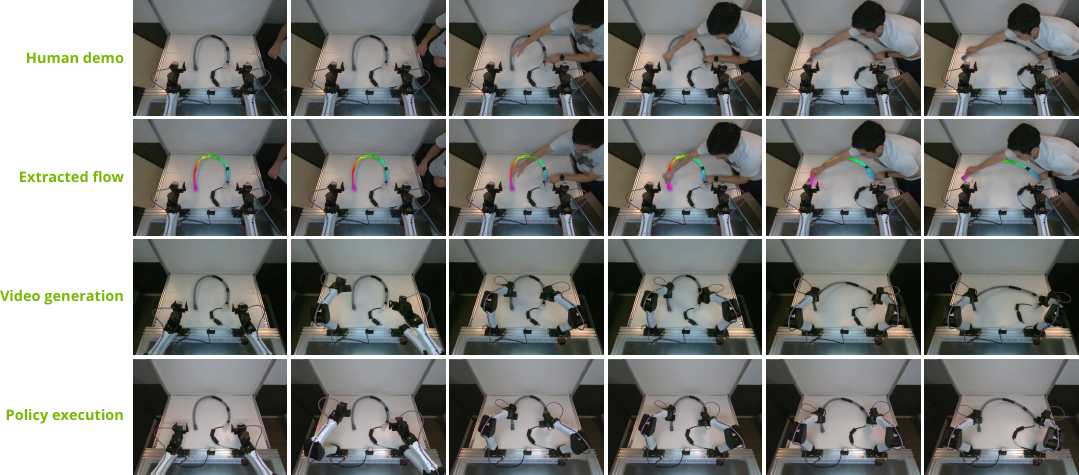}
    \caption{\textbf{Object-Flow-Conditioned Real-Robot Execution.}
    We illustrate the method on a flexible-pipe-bending task.
    The columns show six matched time steps.
    From a held-out human demonstration (row~1), we extract the motion of the manipulated object (row~2) in the form of desired object flow.
    Conditioned on that flow alone, \shortname generates compatible robot motion in row~3.
    A learned action-output head converts the latent rollout into executable robot actions, yielding the successful execution shown in row~4.}
    \label{fig:policy_learning_pipeline}
\end{figure*}

\paragraph{Results.}
As shown in \Cref{fig:policy_evaluation_correlation}, RoboLab policies' generated success rates closely match the reference rates, with Pearson's $r=0.96$, Spearman's $\rho=0.93$, and a mean absolute error of $5.7$ percentage points.
When averaged across tasks, the generated success rates also reproduce the reference ranking of all five policies.

\paragraph{Real-world proof of concept.}
We additionally replay successful and failed cloth-folding trajectories on a real-world setup.
As shown in \Cref{fig:real_world_policy_evaluation}, the generated rollouts qualitatively preserve the different task outcomes, providing preliminary evidence that evaluation can extend beyond the RoboLab benchmark.
\subsection{Flow-Conditioned World Action Model}
\label{sec:policy_learning}

We qualitatively test whether the world action model of \Cref{sec:world_action_model} can convert desired object motion into executable real-robot behavior.

\paragraph{Desired-flow transfer.}
The reported source is a held-out human demonstration.
Only extracted object flow is provided, and embodiment flow is excluded from the input condition, so the conditioned \shortname rollout predicts the required robot motion to drive the desired object flow.
Executable robot actions are produced from the rollout by a trained action-output head.

\paragraph{Real-robot protocol.}
We illustrate the model on one flexible-pipe-bending task (see \Cref{fig:policy_learning_pipeline}).

\section{Related Work}
\subsection{Neural Simulators and Robotic World Models}

Latent world models learn compact dynamics for control and planning by imagining the consequences of candidate actions, as in Dreamer~\citep{hafner_Mastering_2025a}; DINO-WM similarly predicts pretrained visual features to support planning without reconstructing pixels~\citep{zhou_DINOWM_2025}.
Video-generation world models instead synthesize visually inspectable future observations, and video foundation models such as Cosmos 2.5 and 3 extend this capability to world simulation for physical AI~\citep{nvidia_World_2026,nvidia_cosmos_2026}.

Action-conditioned video simulators have been used for policy evaluation and ranking, synthetic policy-improvement data, and interactive rollouts: \cite{quevedo_WorldGym_2025} and \cite{tseng_Scalable_2025} study policy ranking; \cite{guo_CtrlWorld_2025} and \cite{wang_Interactive_2026} also use imagined trajectories for policy improvement or training; and Veo~\cite{team_Evaluating_2026} studies nominal and out-of-distribution policy behavior.
Our focus is a shared visual action interface: action flow maps an executable command to its visible image-plane motion, allowing the same video-prediction interface to support cross-embodiment training and open-loop policy evaluation.

\subsection{Motion-Conditioned Video Generation}

Motion Prompting uses sparse and dense motion trajectories as a flexible video-conditioning signal~\citep{geng_Motion_2024}.
ATI injects user-specified trajectories into video latents through Gaussian propagation, while Wan-Move, our closest video-conditioning precedent, transports first-frame appearance features along latent trajectories to form motion-aware conditions~\citep{wang_ATI_2025a,chu_WanMove_2025}.
Subsequent work has explored forward and inverse motion reasoning with active and passive trajectories, and physical action conditioning through simulated visual intermediates~\citep{liu_MoRight_2026,liu_RealWonder_2026}.

Other approaches use related two-dimensional conditions: MIMIC derives reference-driven semantic masks, Mask2IV models actor and object mask trajectories, BridgeV2W renders command-directed embodiment masks, and OSCAR renders two-dimensional kinematic skeletons~\citep{chen_MIMIC_2025,li_Mask2IV_2025,chen_BridgeV2W_2026,wu_oscar_2026}.
Neighboring geometric and flow interfaces include kinematic pointmap conditioning, optical-flow world-action models, robot rendering, flow-centric generative planning, trajectory-conditioned cross-embodiment skill transfer, and contact-flow conditioning~\citep{xu_Kinema4D_2026,chen_flowwam_2026,kim_robot-factored_2026,gao_FLIP_2025,tang_Trajectory_2025,azirar_contactflow_2026}.
Our approach jointly uses visibility-aware tracks of robot-surface points, multi-embodiment training, and a shared forward/inverse interface, while also recovering training-time tracks from videos that lack robot descriptions or calibration.
Forward deployment still derives action flow from robot geometry and camera calibration.

\section{Limitations}

Qualitatively, the world action model can exhibit centimeter-scale grasp imprecision.
We hypothesize that limited depth awareness may contribute to this imprecision.
Grasp and contact state can also be ambiguous in generated rollouts, including whether an object has been successfully secured.
Future world models could reduce these ambiguities by conditioning on additional modalities such as depth, tactile sensing, and force sensing.
Our wrist-camera experiment is limited to a qualitative DROID proof of concept; systematic evaluation in wrist-camera simulation, mobile manipulation, and settings with broader camera motion remains future work.
Our current evaluation is limited to open-loop policy evaluation; closed-loop evaluation remains an important direction for future work.

\section{Conclusion}

We introduced kinematically grounded action flow as a visual control interface for transferable neural world simulation across robot embodiments and video backbones.
Beyond improving action-conditioned prediction and data-efficient transfer, the interface supports policy evaluation: across RoboLab policies, replayed versus reference success rates have a Pearson correlation of $r=0.96$.
In the inverse direction, our world action model uses desired object flow from a held-out human demonstration to generate compatible robot motion without embodiment-flow input, and a trained action head maps its latent features to executable robot actions without task-specific expert robot demonstrations.
These results position action flow as a shared control interface that separates task specification from execution learning while connecting visual prediction, open-loop policy evaluation, and robot control.

\bibliographystyle{plainnat}  %
\begingroup
\Urlmuskip=0mu plus 1mu\relax
\bibliography{My_Library}  %

\begin{thebibliography}{59}
\providecommand{\natexlab}[1]{#1}
\providecommand{\url}[1]{\texttt{#1}}
\expandafter\ifx\csname urlstyle\endcsname\relax
  \providecommand{\doi}[1]{doi: #1}\else
  \providecommand{\doi}{doi: \begingroup \urlstyle{rm}\Url}\fi

\bibitem[noa(2026)]{noauthor_lightx2v_2026}
{LightX2V}: {Light} {Video} {Generation} {Inference} {Framework}, August 2026.
\newblock URL \url{https://github.com/ModelTC/LightX2V}.
\newblock original-date: 2025-03-24T10:27:56Z.

\bibitem[AgiBot-World-Contributors et~al.(2025)AgiBot-World-Contributors, Bu, Cai, Chen, Cui, Ding, Feng, Gao, He, Hu, Huang, Jiang, Jiang, Jing, Li, Li, Liu, Liu, Lu, Luo, Luo, Mu, Niu, Pan, Pang, Qiao, Ren, Ruan, Shan, Shen, Shi, Shi, Shi, Sima, Song, Wang, Wang, Wei, Xie, Xu, Yan, Yang, Yang, Yang, Yao, Zeng, Zhang, Zhang, Zhao, Zhao, Zhao, and Zhu]{agibot-world-contributors_AgiBot_2025}
AgiBot-World-Contributors, Qingwen Bu, Jisong Cai, Li~Chen, Xiuqi Cui, Yan Ding, Siyuan Feng, Shenyuan Gao, Xindong He, Xuan Hu, Xu~Huang, Shu Jiang, Yuxin Jiang, Cheng Jing, Hongyang Li, Jialu Li, Chiming Liu, Yi~Liu, Yuxiang Lu, Jianlan Luo, Ping Luo, Yao Mu, Yuehan Niu, Yixuan Pan, Jiangmiao Pang, Yu~Qiao, Guanghui Ren, Cheng Ruan, Jiaqi Shan, Yongjian Shen, Chengshi Shi, Mingkang Shi, Modi Shi, Chonghao Sima, Jianheng Song, Huijie Wang, Wenhao Wang, Dafeng Wei, Chengen Xie, Guo Xu, Junchi Yan, Cunbiao Yang, Lei Yang, Shukai Yang, Maoqing Yao, Jia Zeng, Chi Zhang, Qinglin Zhang, Bin Zhao, Chengyue Zhao, Jiaqi Zhao, and Jianchao Zhu.
\newblock {AgiBot} {World} {Colosseo}: {A} {Large}-scale {Manipulation} {Platform} for {Scalable} and {Intelligent} {Embodied} {Systems}, April 2025.
\newblock URL \url{http://arxiv.org/abs/2503.06669}.
\newblock arXiv:2503.06669 [cs].

\bibitem[Allshire et~al.(2026)Allshire, Singh, Singh, Rashid, Choi, McAllister, Yu, Chen, Huang, Abbeel, Chen, Duan, Isola, Malik, Shentu, Shi, Wu, and Kanazawa]{allshire_scalable_2026}
Arthur Allshire, Himanshu~Gaurav Singh, Ritvik Singh, Adam Rashid, Hongsuk Choi, David McAllister, Justin Yu, Yiyuan Chen, Huang Huang, Pieter Abbeel, Xi~Chen, Rocky Duan, Phillip Isola, Jitendra Malik, Fred Shentu, Guanya Shi, Philipp Wu, and Angjoo Kanazawa.
\newblock Scalable {Behavior} {Cloning} with {Open} {Data}, {Training}, and {Evaluation}, June 2026.
\newblock URL \url{http://arxiv.org/abs/2606.27375}.
\newblock arXiv:2606.27375 [cs.RO].

\bibitem[Azirar et~al.(2026)Azirar, Pallotta, Nogga, Gall, Behnke, and Blum]{azirar_contactflow_2026}
Sami Azirar, Enrico Pallotta, Jan Nogga, Jürgen Gall, Sven Behnke, and Hermann Blum.
\newblock {ContactFlow}: {A} video action conditioning that transfers across embodiments, July 2026.
\newblock URL \url{http://arxiv.org/abs/2607.26579}.
\newblock arXiv:2607.26579 [cs.RO].

\bibitem[Bharadhwaj et~al.(2024)Bharadhwaj, Mottaghi, Gupta, and Tulsiani]{bharadhwaj_Track2Act_2024}
Homanga Bharadhwaj, Roozbeh Mottaghi, Abhinav Gupta, and Shubham Tulsiani.
\newblock {Track2Act}: {Predicting} {Point} {Tracks} from {Internet} {Videos} {Enables} {Generalizable} {Robot} {Manipulation}.
\newblock In \emph{European {Conference} on {Computer} {Vision}}, 2024.
\newblock ISBN 978-3-031-73116-7.
\newblock \doi{10.1007/978-3-031-73116-7_18}.

\bibitem[Black et~al.(2024)Black, Brown, Driess, Esmail, Equi, Finn, Fusai, Groom, Hausman, Ichter, Jakubczak, Jones, Ke, Levine, Li-Bell, Mothukuri, Nair, Pertsch, Shi, Tanner, Vuong, Walling, Wang, and Zhilinsky]{black_$pi_0$_2024}
Kevin Black, Noah Brown, Danny Driess, Adnan Esmail, Michael Equi, Chelsea Finn, Niccolo Fusai, Lachy Groom, Karol Hausman, Brian Ichter, Szymon Jakubczak, Tim Jones, Liyiming Ke, Sergey Levine, Adrian Li-Bell, Mohith Mothukuri, Suraj Nair, Karl Pertsch, Lucy~Xiaoyang Shi, James Tanner, Quan Vuong, Anna Walling, Haohuan Wang, and Ury Zhilinsky.
\newblock {\textbackslash}pi\_0: {A} {Vision}-{Language}-{Action} {Flow} {Model} for {General} {Robot} {Control}, November 2024.
\newblock URL \url{http://arxiv.org/abs/2410.24164}.
\newblock arXiv:2410.24164.

\bibitem[Carion et~al.(2025)Carion, Gustafson, Hu, Debnath, Hu, Suris, Ryali, Alwala, Khedr, Huang, Lei, Ma, Guo, Kalla, Marks, Greer, Wang, Sun, Rädle, Afouras, Mavroudi, Xu, Wu, Zhou, Momeni, Hazra, Ding, Vaze, Porcher, Li, Li, Kamath, Cheng, Dollár, Ravi, Saenko, Zhang, and Feichtenhofer]{carion_SAM_2025}
Nicolas Carion, Laura Gustafson, Yuan-Ting Hu, Shoubhik Debnath, Ronghang Hu, Didac Suris, Chaitanya Ryali, Kalyan~Vasudev Alwala, Haitham Khedr, Andrew Huang, Jie Lei, Tengyu Ma, Baishan Guo, Arpit Kalla, Markus Marks, Joseph Greer, Meng Wang, Peize Sun, Roman Rädle, Triantafyllos Afouras, Effrosyni Mavroudi, Katherine Xu, Tsung-Han Wu, Yu~Zhou, Liliane Momeni, Rishi Hazra, Shuangrui Ding, Sagar Vaze, Francois Porcher, Feng Li, Siyuan Li, Aishwarya Kamath, Ho~Kei Cheng, Piotr Dollár, Nikhila Ravi, Kate Saenko, Pengchuan Zhang, and Christoph Feichtenhofer.
\newblock {SAM} 3: {Segment} {Anything} with {Concepts}, November 2025.
\newblock URL \url{http://arxiv.org/abs/2511.16719}.
\newblock arXiv:2511.16719 [cs].

\bibitem[Chen et~al.(2025)Chen, Rong, Chen, Wang, Zhou, Chen, and Ye]{chen_MIMIC_2025}
Tianxiao Chen, Jintao Rong, Huajin Chen, Jingya Wang, Tao Zhou, Jiming Chen, and Qi~Ye.
\newblock {MIMIC}: {Mask}-{Injected} {Manipulation} {Video} {Generation} with {Interaction} {Control}.
\newblock October 2025.
\newblock URL \url{https://openreview.net/forum?id=COrUdVuInH}.

\bibitem[Chen et~al.(2026{\natexlab{a}})Chen, Li, Xu, Ma, Yang, Wang, Yang, An, Guan, Liu, Si, Huang, Liu, Liu, Huang, and Wang]{chen_flowwam_2026}
Yixiang Chen, Peiyan Li, Yuan Xu, Qisen Ma, Jiabing Yang, Kai Wang, Jianhua Yang, Dong An, He~Guan, Gaoteng Liu, Jianlou Si, Jun Huang, Jing Liu, Nianfeng Liu, Yan Huang, and Liang Wang.
\newblock {FlowWAM}: {Optical} {Flow} as a {Unified} {Action} {Representation} for {World} {Action} {Models}, July 2026{\natexlab{a}}.
\newblock URL \url{http://arxiv.org/abs/2607.13017}.
\newblock arXiv:2607.13017 [cs.RO].

\bibitem[Chen et~al.(2026{\natexlab{b}})Chen, Li, Yang, He, Wu, Xu, Wang, Liu, Liu, Huang, and Wang]{chen_BridgeV2W_2026}
Yixiang Chen, Peiyan Li, Jiabing Yang, Keji He, Xiangnan Wu, Yuan Xu, Kai Wang, Jing Liu, Nianfeng Liu, Yan Huang, and Liang Wang.
\newblock {BridgeV2W}: {Bridging} {Video} {Generation} {Models} to {Embodied} {World} {Models} via {Embodiment} {Masks}, February 2026{\natexlab{b}}.
\newblock URL \url{http://arxiv.org/abs/2602.03793}.
\newblock arXiv:2602.03793 [cs].

\bibitem[Chen et~al.(2026{\natexlab{c}})Chen, Wang, Huang, Yang, Zhang, Xiao, Chu, Mao, Hu, Liu, Zhao, Mao, Chen, Xie, Qi, and Han]{chen_longlive-20_2026}
Yukang Chen, Luozhou Wang, Wei Huang, Shuai Yang, Bohan Zhang, Yicheng Xiao, Ruihang Chu, Weian Mao, Qixin Hu, Shaoteng Liu, Yuyang Zhao, Huizi Mao, Ying-Cong Chen, Enze Xie, Xiaojuan Qi, and Song Han.
\newblock {LongLive}-2.0: {An} {NVFP4} {Parallel} {Infrastructure} for {Long} {Video} {Generation}, May 2026{\natexlab{c}}.
\newblock URL \url{http://arxiv.org/abs/2605.18739}.
\newblock arXiv:2605.18739 [cs.CV].

\bibitem[Chi et~al.(2024)Chi, Xu, Pan, Cousineau, Burchfiel, Feng, Tedrake, and Song]{chi_Universal_2024}
Cheng Chi, Zhenjia Xu, Chuer Pan, Eric Cousineau, Benjamin Burchfiel, Siyuan Feng, Russ Tedrake, and Shuran Song.
\newblock Universal {Manipulation} {Interface}: {In}-{The}-{Wild} {Robot} {Teaching} {Without} {In}-{The}-{Wild} {Robots}, February 2024.
\newblock URL \url{http://arxiv.org/abs/2402.10329}.
\newblock arXiv:2402.10329 [cs].

\bibitem[Chu et~al.(2025)Chu, He, Chen, Zhang, Xu, Xia, Wang, Yi, Liu, Zhao, Liu, Zhang, and Yang]{chu_WanMove_2025}
Ruihang Chu, Yefei He, Zhekai Chen, Shiwei Zhang, Xiaogang Xu, Bin Xia, Dingdong Wang, Hongwei Yi, Xihui Liu, Hengshuang Zhao, Yu~Liu, Yingya Zhang, and Yujiu Yang.
\newblock Wan-{Move}: {Motion}-controllable {Video} {Generation} via {Latent} {Trajectory} {Guidance}, December 2025.
\newblock URL \url{http://arxiv.org/abs/2512.08765}.
\newblock arXiv:2512.08765 [cs].

\bibitem[Fang et~al.(2026)Fang, Duan, Clay, Wang, Liu, Huang, Fan, Tsai, Chen, Wang, Xing, Cho, Park, Eftekhar, Sushko, Farley, Wadhwa, Harrison, Han, Lee, VanderBilt, Hendrix, Ellawela, Ngoo, Chai, Ren, Farhadi, Fox, and Krishna]{fang_molmoact2_2026}
Haoquan Fang, Jiafei Duan, Donovan Clay, Sam Wang, Shuo Liu, Weikai Huang, Xiang Fan, Wei-Chuan Tsai, Shirui Chen, Yi~Ru Wang, Shanli Xing, Jaemin Cho, Jae~Sung Park, Ainaz Eftekhar, Peter Sushko, Karen Farley, Angad Wadhwa, Cole Harrison, Winson Han, Ying-Chun Lee, Eli VanderBilt, Rose Hendrix, Suveen Ellawela, Lucas Ngoo, Joyce Chai, Zhongzheng Ren, Ali Farhadi, Dieter Fox, and Ranjay Krishna.
\newblock {MolmoAct2}: {Action} {Reasoning} {Models} for {Real}-world {Deployment}, May 2026.
\newblock URL \url{http://arxiv.org/abs/2605.02881}.
\newblock arXiv:2605.02881 [cs.RO].

\bibitem[Fu et~al.(2026)Fu, Nan, Sun, Li, Qian, Barry, Kitani, and Konidaris]{fu_NovaPlan_2026}
Jiahui Fu, Junyu Nan, Lingfeng Sun, Hongyu Li, Jianing Qian, Jennifer~L. Barry, Kris Kitani, and George Konidaris.
\newblock {NovaPlan}: {Zero}-{Shot} {Long}-{Horizon} {Manipulation} via {Closed}-{Loop} {Video} {Language} {Planning}, February 2026.
\newblock URL \url{http://arxiv.org/abs/2602.20119}.
\newblock arXiv:2602.20119 [cs].

\bibitem[Gao et~al.(2025)Gao, Zhang, Xu, Cai, and Shao]{gao_FLIP_2025}
Chongkai Gao, Haozhuo Zhang, Zhixuan Xu, Zhehao Cai, and Lin Shao.
\newblock {FLIP}: {Flow}-{Centric} {Generative} {Planning} as {General}-{Purpose} {Manipulation} {World} {Model}, February 2025.
\newblock URL \url{http://arxiv.org/abs/2412.08261}.
\newblock arXiv:2412.08261 [cs].

\bibitem[Geng et~al.(2024)Geng, Herrmann, Hur, Cole, Zhang, Pfaff, Lopez-Guevara, Doersch, Aytar, Rubinstein, Sun, Wang, Owens, and Sun]{geng_Motion_2024}
Daniel Geng, Charles Herrmann, Junhwa Hur, Forrester Cole, Serena Zhang, Tobias Pfaff, Tatiana Lopez-Guevara, Carl Doersch, Yusuf Aytar, Michael Rubinstein, Chen Sun, Oliver Wang, Andrew Owens, and Deqing Sun.
\newblock Motion {Prompting}: {Controlling} {Video} {Generation} with {Motion} {Trajectories}, December 2024.
\newblock URL \url{http://arxiv.org/abs/2412.02700}.
\newblock arXiv:2412.02700 [cs].

\bibitem[Gillman et~al.(2026)Gillman, Zhou, Tang, Luo, Chakravarthy, Aggarwal, Freeman, and Sun]{gillman_goal_2026}
Nate Gillman, Yinghua Zhou, Zitian Tang, Evan Luo, Arjan Chakravarthy, Daksh Aggarwal, Michael Freeman, and Chen Sun.
\newblock Goal {Force}: {Teaching} {Video} {Models} {To} {Accomplish} {Physics}-{Conditioned} {Goals}.
\newblock pages 20077--20087, 2026.
\newblock URL \url{https://openaccess.thecvf.com/content/CVPR2026/html/Gillman_Goal_Force_Teaching_Video_Models_To_Accomplish_Physics-Conditioned_Goals_CVPR_2026_paper.html}.

\bibitem[Guo et~al.(2025)Guo, Shi, Chen, and Finn]{guo_CtrlWorld_2025}
Yanjiang Guo, Lucy~Xiaoyang Shi, Jianyu Chen, and Chelsea Finn.
\newblock Ctrl-{World}: {A} {Controllable} {Generative} {World} {Model} for {Robot} {Manipulation}, October 2025.
\newblock URL \url{http://arxiv.org/abs/2510.10125}.
\newblock arXiv:2510.10125 [cs].

\bibitem[Hafner et~al.(2025)Hafner, Pasukonis, Ba, and Lillicrap]{hafner_Mastering_2025a}
Danijar Hafner, Jurgis Pasukonis, Jimmy Ba, and Timothy Lillicrap.
\newblock Mastering diverse control tasks through world models.
\newblock \emph{Nature}, 640\penalty0 (8059):\penalty0 647--653, April 2025.
\newblock ISSN 1476-4687.
\newblock \doi{10.1038/s41586-025-08744-2}.
\newblock URL \url{https://www.nature.com/articles/s41586-025-08744-2}.

\bibitem[Harley et~al.(2025)Harley, You, Sun, Zheng, Raghuraman, Gu, Liang, Chu, Dave, Tokmakov, You, Ambrus, Fragkiadaki, and Guibas]{harley_AllTracker_2025}
Adam~W. Harley, Yang You, Xinglong Sun, Yang Zheng, Nikhil Raghuraman, Yunqi Gu, Sheldon Liang, Wen-Hsuan Chu, Achal Dave, Pavel Tokmakov, Suya You, Rares Ambrus, Katerina Fragkiadaki, and Leonidas~J. Guibas.
\newblock {AllTracker}: {Efficient} {Dense} {Point} {Tracking} at {High} {Resolution}, June 2025.
\newblock URL \url{http://arxiv.org/abs/2506.07310}.
\newblock arXiv:2506.07310 [cs].

\bibitem[Heusel et~al.(2017)Heusel, Ramsauer, Unterthiner, Nessler, and Hochreiter]{heusel_gans_2017}
Martin Heusel, Hubert Ramsauer, Thomas Unterthiner, Bernhard Nessler, and Sepp Hochreiter.
\newblock {GANs} {Trained} by a {Two} {Time}-{Scale} {Update} {Rule} {Converge} to a {Local} {Nash} {Equilibrium}.
\newblock In \emph{Advances in {Neural} {Information} {Processing} {Systems}}, volume~30. Curran Associates, Inc., 2017.
\newblock URL \url{https://proceedings.neurips.cc/paper_files/paper/2017/hash/8a1d694707eb0fefe65871369074926d-Abstract.html}.

\bibitem[Hoque et~al.(2025)Hoque, Huang, Yoon, Sivapurapu, and Zhang]{hoque_EgoDex_2025}
Ryan Hoque, Peide Huang, David~J. Yoon, Mouli Sivapurapu, and Jian Zhang.
\newblock {EgoDex}: {Learning} {Dexterous} {Manipulation} from {Large}-{Scale} {Egocentric} {Video}, May 2025.
\newblock URL \url{http://arxiv.org/abs/2505.11709}.
\newblock arXiv:2505.11709 [cs].

\bibitem[Hu et~al.(2021)Hu, Shen, Wallis, Allen-Zhu, Li, Wang, Wang, and Chen]{hu_lora_2021}
Edward~J. Hu, Yelong Shen, Phillip Wallis, Zeyuan Allen-Zhu, Yuanzhi Li, Shean Wang, Lu~Wang, and Weizhu Chen.
\newblock {LoRA}: {Low}-{Rank} {Adaptation} of {Large} {Language} {Models}, October 2021.
\newblock URL \url{http://arxiv.org/abs/2106.09685}.
\newblock arXiv:2106.09685 [cs.CL].

\bibitem[Intelligence et~al.(2025)Intelligence, Black, Brown, Darpinian, Dhabalia, Driess, Esmail, Equi, Finn, Fusai, Galliker, Ghosh, Groom, Hausman, Ichter, Jakubczak, Jones, Ke, LeBlanc, Levine, Li-Bell, Mothukuri, Nair, Pertsch, Ren, Shi, Smith, Springenberg, Stachowicz, Tanner, Vuong, Walke, Walling, Wang, Yu, and Zhilinsky]{intelligence_$p_05$_2025}
Physical Intelligence, Kevin Black, Noah Brown, James Darpinian, Karan Dhabalia, Danny Driess, Adnan Esmail, Michael Equi, Chelsea Finn, Niccolo Fusai, Manuel~Y. Galliker, Dibya Ghosh, Lachy Groom, Karol Hausman, Brian Ichter, Szymon Jakubczak, Tim Jones, Liyiming Ke, Devin LeBlanc, Sergey Levine, Adrian Li-Bell, Mohith Mothukuri, Suraj Nair, Karl Pertsch, Allen~Z. Ren, Lucy~Xiaoyang Shi, Laura Smith, Jost~Tobias Springenberg, Kyle Stachowicz, James Tanner, Quan Vuong, Homer Walke, Anna Walling, Haohuan Wang, Lili Yu, and Ury Zhilinsky.
\newblock {\textbackslash}pi\_\{0.5\}: a {Vision}-{Language}-{Action} {Model} with {Open}-{World} {Generalization}, April 2025.
\newblock URL \url{http://arxiv.org/abs/2504.16054}.
\newblock arXiv:2504.16054 [cs].

\bibitem[Khazatsky et~al.(2024)Khazatsky, Pertsch, Nair, Balakrishna, Dasari, Karamcheti, Nasiriany, Srirama, Chen, Ellis, Fagan, Hejna, Itkina, Lepert, Ma, Miller, Wu, Belkhale, Dass, Ha, Jain, Lee, Lee, Memmel, Park, Radosavovic, Wang, Zhan, Black, Chi, Hatch, Lin, Lu, Mercat, Rehman, Sanketi, Sharma, Simpson, Vuong, Walke, Wulfe, Xiao, Yang, Yavary, Zhao, Agia, Baijal, Castro, Chen, Chen, Chung, Drake, Foster, Gao, Herrera, Heo, Hsu, Hu, Jackson, Le, Li, Lin, Lin, Ma, Maddukuri, Mirchandani, Morton, Nguyen, O'Neill, Scalise, Seale, Son, Tian, Tran, Wang, Wu, Xie, Yang, Yin, Zhang, Bastani, Berseth, Bohg, Goldberg, Gupta, Gupta, Jayaraman, Lim, Malik, Martín-Martín, Ramamoorthy, Sadigh, Song, Wu, Yip, Zhu, Kollar, Levine, and Finn]{khazatsky_DROID_2024}
Alexander Khazatsky, Karl Pertsch, Suraj Nair, Ashwin Balakrishna, Sudeep Dasari, Siddharth Karamcheti, Soroush Nasiriany, Mohan~Kumar Srirama, Lawrence~Yunliang Chen, Kirsty Ellis, Peter~David Fagan, Joey Hejna, Masha Itkina, Marion Lepert, Yecheng~Jason Ma, Patrick~Tree Miller, Jimmy Wu, Suneel Belkhale, Shivin Dass, Huy Ha, Arhan Jain, Abraham Lee, Youngwoon Lee, Marius Memmel, Sungjae Park, Ilija Radosavovic, Kaiyuan Wang, Albert Zhan, Kevin Black, Cheng Chi, Kyle~Beltran Hatch, Shan Lin, Jingpei Lu, Jean Mercat, Abdul Rehman, Pannag~R. Sanketi, Archit Sharma, Cody Simpson, Quan Vuong, Homer~Rich Walke, Blake Wulfe, Ted Xiao, Jonathan~Heewon Yang, Arefeh Yavary, Tony~Z. Zhao, Christopher Agia, Rohan Baijal, Mateo~Guaman Castro, Daphne Chen, Qiuyu Chen, Trinity Chung, Jaimyn Drake, Ethan~Paul Foster, Jensen Gao, David~Antonio Herrera, Minho Heo, Kyle Hsu, Jiaheng Hu, Donovon Jackson, Charlotte Le, Yunshuang Li, Kevin Lin, Roy Lin, Zehan Ma, Abhiram Maddukuri, Suvir Mirchandani, Daniel Morton, Tony Nguyen, Abigail O'Neill, Rosario Scalise, Derick Seale, Victor Son, Stephen Tian, Emi Tran, Andrew~E. Wang, Yilin Wu, Annie Xie, Jingyun Yang, Patrick Yin, Yunchu Zhang, Osbert Bastani, Glen Berseth, Jeannette Bohg, Ken Goldberg, Abhinav Gupta, Abhishek Gupta, Dinesh Jayaraman, Joseph~J. Lim, Jitendra Malik, Roberto Martín-Martín, Subramanian Ramamoorthy, Dorsa Sadigh, Shuran Song, Jiajun Wu, Michael~C. Yip, Yuke Zhu, Thomas Kollar, Sergey Levine, and Chelsea Finn.
\newblock {DROID}: {A} {Large}-{Scale} {In}-{The}-{Wild} {Robot} {Manipulation} {Dataset}, March 2024.
\newblock URL \url{http://arxiv.org/abs/2403.12945}.
\newblock arXiv:2403.12945 [cs].

\bibitem[Kim et~al.(2026)Kim, Kim, Cha, and Joo]{kim_robot-factored_2026}
Byungjun Kim, Taeksoo Kim, Hyunsoo Cha, and Hanbyul Joo.
\newblock Robot-{Factored} {World} {Models} via {Robot} {Rendering}, July 2026.
\newblock URL \url{http://arxiv.org/abs/2607.22535}.
\newblock arXiv:2607.22535 [cs.RO] version: 1.

\bibitem[Li et~al.(2025{\natexlab{a}})Li, Zhao, Yang, and Sevilla-Lara]{li_Mask2IV_2025}
Gen Li, Bo~Zhao, Jianfei Yang, and Laura Sevilla-Lara.
\newblock {Mask2IV}: {Interaction}-{Centric} {Video} {Generation} via {Mask} {Trajectories}, November 2025{\natexlab{a}}.
\newblock URL \url{http://arxiv.org/abs/2510.03135}.
\newblock arXiv:2510.03135 [cs].

\bibitem[Li et~al.(2025{\natexlab{b}})Li, Sun, Hu, Ta, Barry, Konidaris, and Fu]{li_NovaFlow_2025}
Hongyu Li, Lingfeng Sun, Yafei Hu, Duy Ta, Jennifer Barry, George Konidaris, and Jiahui Fu.
\newblock {NovaFlow}: {Zero}-{Shot} {Manipulation} via {Actionable} {Flow} from {Generated} {Videos}, October 2025{\natexlab{b}}.
\newblock URL \url{http://arxiv.org/abs/2510.08568}.
\newblock arXiv:2510.08568 [cs].

\bibitem[Li et~al.(2026)Li, Fu, Cong, Li, Huang, Jiang, He, Liang, Fu, Lu, Sridhar, Smith, Konidaris, and Li]{li_deform360_2026}
Hongyu Li, Wanjia Fu, Xiaoyan Cong, Zekun Li, Binghao Huang, Hanxiao Jiang, Xintong He, Yiqing Liang, Rao Fu, Tao Lu, Srinath Sridhar, Kevin~A. Smith, George Konidaris, and Yunzhu Li.
\newblock Deform360: {A} {Massive} {Multi}-view {Visuotactile} {Dataset} for {Deformable} {World} {Models}, July 2026.
\newblock URL \url{http://arxiv.org/abs/2607.05390}.
\newblock arXiv:2607.05390 [cs.RO].

\bibitem[Liu et~al.(2026{\natexlab{a}})Liu, Ren, Shen, Ling, Gupta, Wang, Fidler, and Gao]{liu_MoRight_2026}
Shaowei Liu, Xuanchi Ren, Tianchang Shen, Huan Ling, Saurabh Gupta, Shenlong Wang, Sanja Fidler, and Jun Gao.
\newblock {MoRight}: {Motion} {Control} {Done} {Right}, April 2026{\natexlab{a}}.
\newblock URL \url{http://arxiv.org/abs/2604.07348}.
\newblock arXiv:2604.07348 [cs].

\bibitem[Liu et~al.(2026{\natexlab{b}})Liu, Chen, Li, Wang, Yu, and Wu]{liu_RealWonder_2026}
Wei Liu, Ziyu Chen, Zizhang Li, Yue Wang, Hong-Xing Yu, and Jiajun Wu.
\newblock {RealWonder}: {Real}-{Time} {Physical} {Action}-{Conditioned} {Video} {Generation}, March 2026{\natexlab{b}}.
\newblock URL \url{http://arxiv.org/abs/2603.05449}.
\newblock arXiv:2603.05449 [cs].

\bibitem[NVIDIA et~al.(2025{\natexlab{a}})NVIDIA, Bjorck, Castañeda, Cherniadev, Da, Ding, Fan, Fang, Fox, Hu, Huang, Jang, Jiang, Kautz, Kundalia, Lao, Li, Lin, Lin, Liu, Llontop, Magne, Mandlekar, Narayan, Nasiriany, Reed, Tan, Wang, Wang, Wang, Wang, Xiang, Xie, Xu, Xu, Ye, Yu, Zhang, Zhang, Zhao, Zheng, and Zhu]{nvidia_GR00T_2025}
NVIDIA, Johan Bjorck, Fernando Castañeda, Nikita Cherniadev, Xingye Da, Runyu Ding, Linxi~"Jim" Fan, Yu~Fang, Dieter Fox, Fengyuan Hu, Spencer Huang, Joel Jang, Zhenyu Jiang, Jan Kautz, Kaushil Kundalia, Lawrence Lao, Zhiqi Li, Zongyu Lin, Kevin Lin, Guilin Liu, Edith Llontop, Loic Magne, Ajay Mandlekar, Avnish Narayan, Soroush Nasiriany, Scott Reed, You~Liang Tan, Guanzhi Wang, Zu~Wang, Jing Wang, Qi~Wang, Jiannan Xiang, Yuqi Xie, Yinzhen Xu, Zhenjia Xu, Seonghyeon Ye, Zhiding Yu, Ao~Zhang, Hao Zhang, Yizhou Zhao, Ruijie Zheng, and Yuke Zhu.
\newblock {GR00T} {N1}: {An} {Open} {Foundation} {Model} for {Generalist} {Humanoid} {Robots}, March 2025{\natexlab{a}}.
\newblock URL \url{http://arxiv.org/abs/2503.14734}.
\newblock arXiv:2503.14734 [cs].

\bibitem[NVIDIA et~al.(2025{\natexlab{b}})NVIDIA, Mittal, Roth, Tigue, Richard, Zhang, Du, Serrano-Muñoz, Yao, Zurbrügg, Rudin, Wawrzyniak, Rakhsha, Denzler, Heiden, Borovicka, Ahmed, Akinola, Anwar, Carlson, Feng, Garg, Gasoto, Gulich, Guo, Gussert, Hansen, Kulkarni, Li, Liu, Makoviychuk, Malczyk, Mazhar, Moghani, Murali, Noseworthy, Poddubny, Ratliff, Rehberg, Schwarke, Singh, Smith, Tang, Thaker, Trepte, Wyk, Yu, Millane, Ramasamy, Steiner, Subramanian, Volk, Chen, Jawale, Kuruttukulam, Lin, Mandlekar, Patzwaldt, Welsh, Zhao, Anes, Lafleche, Moënne-Loccoz, Park, Stepinski, Gelder, Amevor, Carius, Chang, Chen, Ciechomski, Daviet, Mohajerani, Muralt, Reutskyy, Sauter, Schirm, Shi, Terdiman, Vilella, Widmer, Yeoman, Chen, Grizan, Li, Li, Smith, Wiltz, Alexis, Chang, Chu, Fan, Farshidian, Handa, Huang, Hutter, Narang, Pouya, Sheng, Zhu, Macklin, Moravanszky, Reist, Guo, Hoeller, and State]{nvidia_isaac_2025}
NVIDIA, Mayank Mittal, Pascal Roth, James Tigue, Antoine Richard, Octi Zhang, Peter Du, Antonio Serrano-Muñoz, Xinjie Yao, René Zurbrügg, Nikita Rudin, Lukasz Wawrzyniak, Milad Rakhsha, Alain Denzler, Eric Heiden, Ales Borovicka, Ossama Ahmed, Iretiayo Akinola, Abrar Anwar, Mark~T. Carlson, Ji~Yuan Feng, Animesh Garg, Renato Gasoto, Lionel Gulich, Yijie Guo, M.~Gussert, Alex Hansen, Mihir Kulkarni, Chenran Li, Wei Liu, Viktor Makoviychuk, Grzegorz Malczyk, Hammad Mazhar, Masoud Moghani, Adithyavairavan Murali, Michael Noseworthy, Alexander Poddubny, Nathan Ratliff, Welf Rehberg, Clemens Schwarke, Ritvik Singh, James~Latham Smith, Bingjie Tang, Ruchik Thaker, Matthew Trepte, Karl~Van Wyk, Fangzhou Yu, Alex Millane, Vikram Ramasamy, Remo Steiner, Sangeeta Subramanian, Clemens Volk, C.~Y. Chen, Neel Jawale, Ashwin~Varghese Kuruttukulam, Michael~A. Lin, Ajay Mandlekar, Karsten Patzwaldt, John Welsh, Huihua Zhao, Fatima Anes, Jean-Francois Lafleche, Nicolas Moënne-Loccoz, Soowan Park, Rob Stepinski, Dirk~Van Gelder, Chris Amevor, Jan Carius, Jumyung Chang, Anka~He Chen, Pablo de~Heras Ciechomski, Gilles Daviet, Mohammad Mohajerani, Julia~von Muralt, Viktor Reutskyy, Michael Sauter, Simon Schirm, Eric~L. Shi, Pierre Terdiman, Kenny Vilella, Tobias Widmer, Gordon Yeoman, Tiffany Chen, Sergey Grizan, Cathy Li, Lotus Li, Connor Smith, Rafael Wiltz, Kostas Alexis, Yan Chang, David Chu, Linxi~"Jim" Fan, Farbod Farshidian, Ankur Handa, Spencer Huang, Marco Hutter, Yashraj Narang, Soha Pouya, Shiwei Sheng, Yuke Zhu, Miles Macklin, Adam Moravanszky, Philipp Reist, Yunrong Guo, David Hoeller, and Gavriel State.
\newblock Isaac {Lab}: {A} {GPU}-{Accelerated} {Simulation} {Framework} for {Multi}-{Modal} {Robot} {Learning}, November 2025{\natexlab{b}}.
\newblock URL \url{http://arxiv.org/abs/2511.04831}.
\newblock arXiv:2511.04831 [cs.RO].

\bibitem[NVIDIA et~al.(2026{\natexlab{a}})NVIDIA, Aditi, Agarwal, Ali, Allen, Antolini, Aubame, Azzolini, Bai, Bala, Balaji, Bapst, Basant, Beladiya, Bhat, Bhat, Blick, Brighella, Cai, Cai, Cameracci, Cao, Cao, Carlson, Casanova, Chang, Chang, Chao, Chattopadhyay, Chaudhari, Chen, Chen, Chen, Chen, Chen, Chen, Chen, Cheng, Cheng, Chia, Choi, Chung, Cong, Cui, Dadela, Dadhich, Dai, Daw, Degirmenci, Monte, Denomme, Dharur, Lucca, Ding, Ding, Ding, Dong, Drumheller, Du, Dzhumamuratova, Efitorov, Eghbalzadeh, Eigbe, Hanafi, Eslami, Falk, Fan, Fan, Fasale, Fefilatyev, Feng, Ferroni, Fidler, Fu, Fugro, Gaikwad, Galda, Gao, Gao, Ge, Ghosh, Goel, Goel, Gokul, Govindaraju, Gu, Guerrero, Guo, Gupta, Gururani, Hadfield, Han, Handa, Hao, Harrim, Hassani, Hayes-Roth, He, Helvig, Hogg, Huang, Huang, Huang, Huang, Huffman, Hutchins, Indupuru, Ivanovic, Jain, Jang, Ji, Jian, Jiang, Jin, Joshi, Joshi, Joshi, Ju, Jung, Kang, Kassekert, Kautz, Khetan, Kiczka, Kierat, Kim, Kim, Kim, Kong, Kong, Kong, Kornuta, Krivov, Kuang, Kumar, Kuo, Kurian, Kutak, Lafleche, Lahkar, Laymoun, Lee, Lee, Leone, Li, Li, Li, Li, Li, Li, Li, Li, Li, Li, Li, Li, Liang, Liao, Lin, Lin, Liu, Liu, Liu, Lu, Lu, Luo, Luo, Luo, Lyu, Ma, Ma, Ma, Majchrowski, Marcoux, Martin, Miao, Mirzaei, Misra, Mo, Mohsin, Moon, Morkisz, Motiian, Motkov, Nah, Narang, Narayanan, Ngazimbi, Ouyang, Pachori, Page, Pang, Park, Patekar, Patwary, Pavone, Pham, Ping, Pouya, Prabhumoye, Praveen, Qu, Rabeti, Ramezanali, Reeb, Ren, Rumley, Rymer, Saito, Seol, Shao, Shekdar, Shen, Shi, Shi, Shi, Shih, Shoeybi, Sieniawski, Song, Sotelo, Sotoodeh, Srinivasa, Srinivasakumar, Stefaniak, Steiger, Sun, Tang, Tang, Tang, Tang, Tavakkoli, Ting, Tomala, Tseng, Varghese, Vasilev, Volk, Wagwani, Waleffe, Wang, Wang, Wang, Wang, Wang, Wang, Wang, Wang, Wang, Watve, Wehr, Wei, Weng, Wu, Wu, Xia, Xiao, Xiao, Xie, Xu, Xu, Xu, Xu, Xu, Xu, Yang, Yang, Yang, Yang, Yang, Yang, You, Yu, Yuan, Yuen, Zeng, Zeren, Zha, Zhang, Zhang, Zhang, Zhang, Zhang, Zhang, Zhang, Zhang, Zhao, Zhao, Zhautouskaya, Zhou, Zhou, Zhu, Zhu, Zhylko, and Zolkowski]{nvidia_cosmos_2026}
NVIDIA, Aditi, Niket Agarwal, Arslan Ali, Jon Allen, Martin Antolini, Adeline Aubame, Alisson Azzolini, Junjie Bai, Maciej Bala, Yogesh Balaji, Josh Bapst, Aarti Basant, Mukesh Beladiya, Mohammad~Qazim Bhat, Zaid~Pervaiz Bhat, Dan Blick, Vanni Brighella, Han Cai, Tiffany Cai, Eric Cameracci, Jiaxin Cao, Yulong Cao, Mark Carlson, Carlos Casanova, Ting-Yun Chang, Yan Chang, Yu-Wei Chao, Prithvijit Chattopadhyay, Roshan Chaudhari, Chieh-Yun Chen, Junyu Chen, Ke~Chen, Qizhi Chen, Wenkai Chen, Xiaotong Chen, Yu~Chen, An-Chieh Cheng, Click Cheng, Xiu Chia, Jeana Choi, Chaeyeon Chung, Wenyan Cong, Yin Cui, Magdalena Dadela, Nalin Dadhich, Wenliang Dai, Joyjit Daw, Alperen Degirmenci, Rodrigo Vieira~Del Monte, Robert Denomme, Sameer Dharur, Marco~Di Lucca, Ke~Ding, Wenhao Ding, Yifan Ding, Yuzhu Dong, Nicole Drumheller, Yilun Du, Aigul Dzhumamuratova, Aleksandr Efitorov, Hamid Eghbalzadeh, Naomi Eigbe, Imad~El Hanafi, Hassan Eslami, Benedikt Falk, Jiaojiao Fan, Jim Fan, Amol Fasale, Sergiy Fefilatyev, Liang Feng, Francesco Ferroni, Sanja Fidler, Xiao Fu, Vikram Fugro, Prashant Gaikwad, T.~J. Galda, Katelyn Gao, Yihuai Gao, Wenhang Ge, Sreyan Ghosh, Arushi Goel, Vivek Goel, Akash Gokul, Rama Govindaraju, Jinwei Gu, Miguel Guerrero, Elfie Guo, Aryaman Gupta, Siddharth Gururani, Hugo Hadfield, Song Han, Ankur Handa, Zekun Hao, Mohammad Harrim, Ali Hassani, Nathan Hayes-Roth, Yufan He, Chris Helvig, Cyrus Hogg, Madison Huang, Michael Huang, Sophia Huang, Yufan Huang, Jacob Huffman, DeLesley Hutchins, Suneel Indupuru, Boris Ivanovic, Arihant Jain, Joel Jang, Ryan Ji, Yanan Jian, Dongfu Jiang, Jingyi Jin, Atharva Joshi, Nikhilesh Joshi, Pranjali Joshi, Andy Ju, Jaehun Jung, Weiwei Kang, Scott Kassekert, Jan Kautz, Ashna Khetan, Julia Kiczka, Slawek Kierat, Gwanghyun Kim, Kuno Kim, Sunny Kim, Kezhi Kong, Xin Kong, Zhifeng Kong, Tomasz Kornuta, Egor Krivov, Hui Kuang, Saurav Kumar, Chia-Wen Kuo, George Kurian, Wojciech Kutak, J.~F. Lafleche, Himangshu Lahkar, Omar Laymoun, Jayjun Lee, Sanggil Lee, Gabriele Leone, Boyi Li, Freya Li, Jiajun Li, Jinfeng Li, Ling Li, Pengcheng Li, Shangru Li, Tingle Li, Xiaolong Li, Xuan Li, Zhaoshuo Li, Zhiqi Li, Hao Liang, Maosheng Liao, Chen-Hsuan Lin, Tsung-Yi Lin, Ming-Yu Liu, Sifei Liu, Zihan Liu, Hai~Loc Lu, Xiangyu Lu, Alice Luo, Ruipu Luo, Wenjie Luo, Jiangran Lyu, Martin~Ding Ma, Nic Ma, Qianli Ma, Dawid Majchrowski, Louis Marcoux, Miguel Martin, Qing Miao, Ashkan Mirzaei, Shreyas Misra, Kaichun Mo, Durra Mohsin, Hyejin Moon, Pawel Morkisz, Saeid Motiian, Kirill Motkov, Seungjun Nah, Yashraj Narang, Deepak Narayanan, Thabang Ngazimbi, Julian Ouyang, Shubham Pachori, David Page, Yatian Pang, Sehwi Park, Mahesh Patekar, Mostofa Patwary, Marco Pavone, Trung Pham, Wei Ping, Soha Pouya, Shrimai Prabhumoye, Varun Praveen, Delin Qu, Hesam Rabeti, Morteza Ramezanali, Marilyn Reeb, Xuanchi Ren, Kristen Rumley, Wojciech Rymer, Jun Saito, Yeongho Seol, John Shao, Piyush Shekdar, Tianwei Shen, Humphrey Shi, Min Shi, Stella Shi, Kevin Shih, Mohammad Shoeybi, Mateusz Sieniawski, Shuran Song, Alexander Sotelo, Amir Sotoodeh, Sunil Srinivasa, Vignesh Srinivasakumar, Bartosz Stefaniak, Rahul~Heinrich Steiger, Shangkun Sun, Jiaxiang Tang, Shitao Tang, Yangyang Tang, Yue Tang, Tolou Tavakkoli, Kayley Ting, Krzysztof Tomala, Wei-Cheng Tseng, Jibin Varghese, Sergei Vasilev, Thomas Volk, Raju Wagwani, Roger Waleffe, Andrew~Z. Wang, Boxiang Wang, Haoxiang Wang, Qiao Wang, Shihao Wang, Shijie Wang, Ting-Chun Wang, Yan Wang, Yu~Wang, Rohit Watve, David Wehr, Fangyin Wei, Xinshuo Weng, Jay~Zhangjie Wu, Kedi Wu, Hongchi Xia, Summer Xiao, Tianjun Xiao, Kevin Xie, Daguang Xu, Jiashu Xu, Mengyao Xu, Ruqing Xu, Xingqian Xu, Yao Xu, Dinghao Yang, Dong Yang, Hans Yang, Xiaodong Yang, Xuning Yang, Yichu Yang, Yurong You, Zhiding Yu, Hao Yuan, Simon Yuen, Xiaohui Zeng, Pengcuo Zeren, Cindy Zha, Haotian Zhang, Jenny Zhang, Jing Zhang, Liangkai Zhang, Paris Zhang, Shun Zhang, Xuanmeng Zhang, Zhizheng Zhang, Ann Zhao, Yilin Zhao, Yuliya Zhautouskaya, Charles Zhou, Fengzhe Zhou, Shilin Zhu, Yuke Zhu, Dima Zhylko, and Artur Zolkowski.
\newblock Cosmos 3: {Omnimodal} {World} {Models} for {Physical} {AI}, June 2026{\natexlab{a}}.
\newblock URL \url{http://arxiv.org/abs/2606.02800}.
\newblock arXiv:2606.02800 [cs.CV].

\bibitem[NVIDIA et~al.(2026{\natexlab{b}})NVIDIA, Ali, Bai, Bala, Balaji, Blakeman, Cai, Cao, Cao, Cha, Chao, Chattopadhyay, Chen, Chen, Chen, Cheng, Cui, Diamond, Ding, Fan, Fan, Feng, Ferroni, Fidler, Fu, Gao, Ge, Gu, Gupta, Gururani, Hanafi, Hassani, Hao, Huffman, Jang, Jannaty, Kautz, Lam, Li, Li, Liao, Lin, Lin, Lin, Ling, Liu, Liu, Lu, Luo, Ma, Mao, Mo, Nah, Narang, Panaskar, Pavao, Pham, Ramezanali, Reda, Reed, Ren, Shao, Shen, Shi, Song, Stefaniak, Sun, Tang, Tasmeen, Tchapmi, Tseng, Varghese, Wang, Wang, Wang, Wang, Wang, Wei, Xu, Yang, Yang, Ye, Ye, Zeng, Zhang, Zhang, Zheng, Zhu, and Zhu]{nvidia_World_2026}
NVIDIA, Arslan Ali, Junjie Bai, Maciej Bala, Yogesh Balaji, Aaron Blakeman, Tiffany Cai, Jiaxin Cao, Tianshi Cao, Elizabeth Cha, Yu-Wei Chao, Prithvijit Chattopadhyay, Mike Chen, Yongxin Chen, Yu~Chen, Shuai Cheng, Yin Cui, Jenna Diamond, Yifan Ding, Jiaojiao Fan, Linxi Fan, Liang Feng, Francesco Ferroni, Sanja Fidler, Xiao Fu, Ruiyuan Gao, Yunhao Ge, Jinwei Gu, Aryaman Gupta, Siddharth Gururani, Imad~El Hanafi, Ali Hassani, Zekun Hao, Jacob Huffman, Joel Jang, Pooya Jannaty, Jan Kautz, Grace Lam, Xuan Li, Zhaoshuo Li, Maosheng Liao, Chen-Hsuan Lin, Tsung-Yi Lin, Yen-Chen Lin, Huan Ling, Ming-Yu Liu, Xian Liu, Yifan Lu, Alice Luo, Qianli Ma, Hanzi Mao, Kaichun Mo, Seungjun Nah, Yashraj Narang, Abhijeet Panaskar, Lindsey Pavao, Trung Pham, Morteza Ramezanali, Fitsum Reda, Scott Reed, Xuanchi Ren, Haonan Shao, Yue Shen, Stella Shi, Shuran Song, Bartosz Stefaniak, Shangkun Sun, Shitao Tang, Sameena Tasmeen, Lyne Tchapmi, Wei-Cheng Tseng, Jibin Varghese, Andrew~Z. Wang, Hao Wang, Haoxiang Wang, Heng Wang, Ting-Chun Wang, Fangyin Wei, Jiashu Xu, Dinghao Yang, Xiaodong Yang, Haotian Ye, Seonghyeon Ye, Xiaohui Zeng, Jing Zhang, Qinsheng Zhang, Kaiwen Zheng, Andrew Zhu, and Yuke Zhu.
\newblock World {Simulation} with {Video} {Foundation} {Models} for {Physical} {AI}, February 2026{\natexlab{b}}.
\newblock URL \url{http://arxiv.org/abs/2511.00062}.
\newblock arXiv:2511.00062 [cs].

\bibitem[Quevedo et~al.(2025)Quevedo, Sharma, Sun, Suryavanshi, Liang, and Yang]{quevedo_WorldGym_2025}
Julian~Hector Quevedo, Ansh~Kumar Sharma, Yixiang Sun, Varad Suryavanshi, Percy Liang, and Sherry Yang.
\newblock {WorldGym}: {World} {Model} as {An} {Environment} for {Policy} {Evaluation}.
\newblock October 2025.
\newblock URL \url{https://openreview.net/forum?id=hidBHy1CAw}.

\bibitem[Tang et~al.(2025)Tang, Lou, Han, Song, Ye, Wang, and Zhao]{tang_Trajectory_2025}
YuHang Tang, Yixuan Lou, Pengfei Han, Haoming Song, Xinyi Ye, Dong Wang, and Bin Zhao.
\newblock Trajectory {Conditioned} {Cross}-embodiment {Skill} {Transfer}, October 2025.
\newblock URL \url{http://arxiv.org/abs/2510.07773}.
\newblock arXiv:2510.07773 [cs].

\bibitem[Team et~al.(2026{\natexlab{a}})Team, Choromanski, Devin, Du, Dwibedi, Gao, Jindal, Kipf, Kirmani, Leal, Liu, Majumdar, Marmon, Parada, Rubanova, Shah, Sindhwani, Tan, Xia, Xiao, Yang, Yu, and Zhou]{team_Evaluating_2026}
Gemini~Robotics Team, Krzysztof Choromanski, Coline Devin, Yilun Du, Debidatta Dwibedi, Ruiqi Gao, Abhishek Jindal, Thomas Kipf, Sean Kirmani, Isabel Leal, Fangchen Liu, Anirudha Majumdar, Andrew Marmon, Carolina Parada, Yulia Rubanova, Dhruv Shah, Vikas Sindhwani, Jie Tan, Fei Xia, Ted Xiao, Sherry Yang, Wenhao Yu, and Allan Zhou.
\newblock Evaluating {Gemini} {Robotics} {Policies} in a {Veo} {World} {Simulator}, January 2026{\natexlab{a}}.
\newblock URL \url{http://arxiv.org/abs/2512.10675}.
\newblock arXiv:2512.10675 [cs].

\bibitem[Team et~al.(2026{\natexlab{b}})Team, Abd, Aggarwal, Algayres, Andreev, Bachem, Ballantyne, Brick, Cărbune, Casbon, Chaturvedi, Chawla, Cotruta, Coucke, Culliton, Dadashi, Dixon, Elhawaty, Evci, Farabet, Ferret, Galgani, Girgin, Grill, Grootendorst, Guo, Hardin, He, Hernandez, Homburger, Hussenot, Ji, Joulin, Kamath, Kassraie, Lacombe, Lahoti, Liu, Martins, Martins, Matejovicova, Merhej, Momchev, Mondal, Mullins, Panyam, Pathak, Perrin, Pinto, Pot, Pouget, Ramé, Ramos, Reid, Rim, Rivière, Roth, Rouillard, Sanseviero, Sessa, Settle, Sinopalnikov, Smoot, Stanczyk, Steiner, Stewart, Tolstikhin, Tschannen, Tsitsulin, Vieillard, Wu, Xu, Yang, Yvinec, Zhang, Zhang, Zou, Aagnes, Abdelhamed, Adamek, Agrawal, Agrawal, Alabdulmohsin, Alayrac, Alon, Amarnath, Anand, Anastasiou, Ariafar, Aubet, Axiotis, Barbero, Barral, Bendebury, Bergmann, Bileschi, Black, Blondel, Borgeaud, Bražinskas, Burnell, Busa-Fekete, Cai, Calandriello, Cameron, Caucheteux, Chaabouni, Chadha, Chan, Chen, Chen, Chen, Chen, Cheng, Chien, Chinaev, Chou, Chu, Coleman, Consul, Conway-Rahman, Crowell, Cutler, Dani, Daruki, Das, Deutsch, Dikkala, Ding, Ding, Dodhia, Donhauser, Doshi, Dragan, Druinsky, Dua, Egyed, Eisenbud, Eppens, Fan, Fatemi, Fathullah, Feinberg, Ferev, Flennerhag, Fujimoto, Oliveira, Galatzer-Levy, Gante, Geisler, Ghosal, Girgis, Glehn, Go, Gokhale, Grills, Gu, Gupta, Gupta, Guruganesh, Hadsell, Harkous, Harlalka, Hassabis, Hauth, Heyward, Hosseini, Hsia, Hsu, Huang, Huang, Hui, Hutter, I, Iliopoulos, Jain, Jawahar, Ji, Jin, Johnson, Joshi, Kandoor, Kang, Kavukcuoglu, Kazemi, Kenealy, Khalifa, Kirk, Korotkov, Kothawade, Kovalev, Kovelamudi, Kraft, Kumar, Kumar, Kuppam, Lannin, Lee, Lee, Lepikhin, Levkovitch, Li, Li, Liévin, Lin, Lin, Liu, Liu, Liu, Liu, Lobov, Lunayach, Ma, Madan, Maksai, Malmi, Matuszak, McDuff, Menghani, Mikuła, Mirylenka, Misiunas, Misra, Mitran, Mohamed, Mukha, Noland, O'Donnell, O'Donoghue, Olszewska, Orlando, Pan, Panigrahy, Parekh, Perez-Nieves, Park, Paskie, Peng, Petrini, Petrov, Pfeiffer, Piot, Plomecka, Poder, Ponce, Pramanik, Racz, Rajan, Ramanovich, Rao, Ritter, Rodrigues, Rosen, Rybiński, Sachdeva, Sander, Sathyanarayana, Savla, Schmidgall, Schuster, Scrivener, Seguin, Sellergren, Severyn, Shafran, Shah, Shahriari, Shangguan, Shenoy, Shenoy, Shivanna, Sho, Spangher, Stokowiec, Strother, Su, Sun, Sundararajan, Tacchetti, Taege, Tafti, Tarbouriech, Tekur, Thakoor, Thapa, Traverse, Treven, Tu, Tung, Ünlü, Veličković, Venkat, Venkatesh, Venkiteswaran, Visin, Vitvitskyi, Vodrahalli, Wang, Wang, Warkentin, Wassenberg, Wieting, Wu, Xiao, Xu, Xu, Xue, Yadav, Yan, Yang, Yang, Yang, Ying, Yoo, Zadimoghaddam, Zafar, Zhang, Zhang, Zhang, Zhang, Zhao, Zhou, and Zou]{team_gemma_2026}
Gemma Team, Sherif~El Abd, Vaibhav Aggarwal, Robin Algayres, Alek Andreev, Olivier Bachem, Ian Ballantyne, Cormac Brick, Victor Cărbune, Michelle Casbon, Mayank Chaturvedi, Aditya Chawla, Victor Cotruta, Alice Coucke, Phil Culliton, Robert Dadashi, Lucas Dixon, Mohamed Elhawaty, Utku Evci, Clément Farabet, Johan Ferret, Filippo Galgani, Sertan Girgin, Jean-Bastien Grill, Maarten Grootendorst, Jiaxian Guo, Cassidy Hardin, Yanzhang He, Steven~M. Hernandez, Omri Homburger, Léonard Hussenot, Juyeong Ji, Armand Joulin, Aishwarya Kamath, Parnian Kassraie, Olivier Lacombe, Preethi Lahoti, Gaël Liu, Gus Martins, Luciano Martins, Tatiana Matejovicova, Ramona Merhej, Nikola Momchev, Sneha Mondal, Ryan Mullins, Sindhu~Raghuram Panyam, Shreya Pathak, Sarah Perrin, André~Susano Pinto, Etienne Pot, Angéline Pouget, Alexandre Ramé, Sabela Ramos, Douglas Reid, David Rim, Morgane Rivière, Karsten Roth, Louis Rouillard, Omar Sanseviero, Pier~Giuseppe Sessa, Shane Settle, Danila Sinopalnikov, Sara Smoot, Piotr Stanczyk, Andreas Steiner, Lawrence Stewart, Ilya Tolstikhin, Michael Tschannen, Anton Tsitsulin, Nino Vieillard, Renjie Wu, Pingmei Xu, Haichuan Yang, Edouard Yvinec, Biao Zhang, Li~Zhang, Joe Zou, Nicolas Aagnes, Abdelrahman Abdelhamed, Jakub Adamek, Shivani Agrawal, Shubham Agrawal, Ibrahim Alabdulmohsin, Jean~Baptiste Alayrac, Uri Alon, Chandramouli Amarnath, Ankesh Anand, Chrysovalantis Anastasiou, Setareh Ariafar, François-Xavier Aubet, Kyriakos Axiotis, Federico Barbero, Joelle Barral, Alexei Bendebury, Urs Bergmann, Stanley Bileschi, Kat Black, Mathieu Blondel, Sebastian Borgeaud, Arthur Bražinskas, Ryan Burnell, Robert Busa-Fekete, Mu~Cai, Daniele Calandriello, Glenn Cameron, Charlotte Caucheteux, Rahma Chaabouni, Garima Chadha, Jetha Chan, Blake~Jianhang Chen, Jesse Chen, Lin Chen, Xu~Chen, Derek Cheng, Tzu-hsiang Chien, Nikolai Chinaev, Yi~Chou, Zhaohui Chu, Benjamin Coleman, Pooja Consul, Sam Conway-Rahman, Scott Crowell, Dylan Cutler, Vivek Dani, Samira Daruki, Anil Das, Daniel Deutsch, Nishanth Dikkala, Li~Ding, Qiuhan Ding, Shenil Dodhia, Konstantin Donhauser, Tulsee Doshi, Anca Dragan, Alex Druinsky, Sahil Dua, Zoltan Egyed, Danielle Eisenbud, Daniel Eppens, Cindy Fan, Bahare Fatemi, Yassir Fathullah, Vlad Feinberg, Milen Ferev, Sebastian Flennerhag, Takumi Fujimoto, João~Gabriel Oliveira, Isaac Galatzer-Levy, João Gante, Simon Geisler, Soham Ghosal, Antonious~M. Girgis, Tamara~von Glehn, Alec Go, Alhaad Gokhale, Alex Grills, Yiming Gu, Mayank Gupta, Pramod Gupta, Guru Guruganesh, Raia Hadsell, Hamza Harkous, Jitendra Harlalka, Demis Hassabis, Anja Hauth, Joe Heyward, Arian Hosseini, Chih-Yang Hsia, I.-Hung Hsu, Xiaopeng Huang, Yangsibo Huang, Kevin Hui, Adrian Hutter, Te~I, Fotis Iliopoulos, Advait Jain, Ganesh Jawahar, Ziwei Ji, Qilin Jin, Melvin Johnson, Kandarp Joshi, Arun Kandoor, Wang-Cheng Kang, Koray Kavukcuoglu, Mehran Kazemi, Kathleen Kenealy, Amr Khalifa, Phoebe Kirk, Ivan Korotkov, Suraj Kothawade, Vitaly Kovalev, Neel Kovelamudi, Adam Kraft, Ravin Kumar, Vivek Kumar, Harish Kuppam, Justin Lannin, Chen-Yu Lee, Seungji Lee, Dmitry Lepikhin, Alon Levkovitch, Dongdong Li, Qiujia Li, Valentin Liévin, Ethan Lin, Ziqian Lin, Casper Liu, Tianlin Liu, Tianqi Liu, Xin Liu, Ivan Lobov, Mayank Lunayach, Min Ma, Gagan Madan, Andrii Maksai, Eric Malmi, Michal Matuszak, Daniel McDuff, Gaurav Menghani, Maciej Mikuła, Daniil Mirylenka, Karolis Misiunas, Vedant Misra, Andreea Mitran, Kareem Mohamed, Maksim Mukha, Eric Noland, James O'Donnell, Brendan O'Donoghue, Kate Olszewska, Bernett Orlando, Wanqiong Pan, Rina Panigrahy, Unnati Parekh, Nicolas Perez-Nieves, Chunjong Park, Eric Paskie, Liqian Peng, Bryce Petrini, Slav Petrov, Jonas Pfeiffer, Bilal Piot, Martyna Plomecka, Siim Poder, Octavio Ponce, Arijit Pramanik, David Racz, Anish Rajan, Michelle Ramanovich, Anand Rao, Marvin Ritter, Vitor Rodrigues, Evan Rosen, Mikołaj Rybiński, Noveen Sachdeva, Michaël~E. Sander, Rohit Sathyanarayana, Sagar Savla, Samuel Schmidgall, Tal Schuster, George Scrivener, Benoit Seguin, Andrew Sellergren, Aliaksei Severyn, Izhak Shafran, Dhruv Shah, Bobak Shahriari, Yuan Shangguan, Ashish Shenoy, Pradeep Shenoy, Rakesh Shivanna, Pauline Sho, Lucas Spangher, Wojciech Stokowiec, Tim Strother, Yao Su, Yinghao Sun, Mukund Sundararajan, Andrea Tacchetti, Mor~Hazan Taege, Pouya Tafti, Jean Tarbouriech, Chetan Tekur, Shantanu Thakoor, Rahul Thapa, Madeleine Traverse, Lenart Treven, Tao Tu, Chien~Te Tung, Çağlar Ünlü, Petar Veličković, Malini~Pooni Venkat, Sagar~Gubbi Venkatesh, Vidya Venkiteswaran, Francesco Visin, Alex Vitvitskyi, Kiran Vodrahalli, Weiyi Wang, Xin Wang, Tris Warkentin, Jan Wassenberg, John Wieting, Cindy Wu, Lechao Xiao, Hao Xu, Yuhui Xu, Fuzhao Xue, Arun Yadav, Jun Yan, Antoine Yang, Lin Yang, Ming-Hsuan Yang, Ziyu Ying, Jae~Hyeon Yoo, Morteza Zadimoghaddam, Sajjad Zafar, Fred Zhang, Jiageng Zhang, Jianyi Zhang, Xiaofan Zhang, Chao Zhao, David Zhou, and Chen Zou.
\newblock Gemma 4 {Technical} {Report}, July 2026{\natexlab{b}}.
\newblock URL \url{http://arxiv.org/abs/2607.02770}.
\newblock arXiv:2607.02770 [cs.CL].

\bibitem[Tseng et~al.(2025)Tseng, Gu, Zhang, Mao, Liu, Shkurti, and Yen-Chen]{tseng_Scalable_2025}
Wei-Cheng Tseng, Jinwei Gu, Qinsheng Zhang, Hanzi Mao, Ming-Yu Liu, Florian Shkurti, and Lin Yen-Chen.
\newblock Scalable {Policy} {Evaluation} with {Video} {World} {Models}, December 2025.
\newblock URL \url{http://arxiv.org/abs/2511.11520}.
\newblock arXiv:2511.11520 [cs].

\bibitem[Unterthiner et~al.(2019)Unterthiner, Steenkiste, Kurach, Marinier, Michalski, and Gelly]{unterthiner_towards_2019}
Thomas Unterthiner, Sjoerd~van Steenkiste, Karol Kurach, Raphael Marinier, Marcin Michalski, and Sylvain Gelly.
\newblock Towards {Accurate} {Generative} {Models} of {Video}: {A} {New} {Metric} \& {Challenges}, March 2019.
\newblock URL \url{http://arxiv.org/abs/1812.01717}.
\newblock arXiv:1812.01717 [cs.CV].

\bibitem[Walke et~al.(2023)Walke, Black, Zhao, Vuong, Zheng, Hansen-Estruch, He, Myers, Kim, Du, Lee, Fang, Finn, and Levine]{walke_BridgeData_2023}
Homer~Rich Walke, Kevin Black, Tony~Z. Zhao, Quan Vuong, Chongyi Zheng, Philippe Hansen-Estruch, Andre~Wang He, Vivek Myers, Moo~Jin Kim, Max Du, Abraham Lee, Kuan Fang, Chelsea Finn, and Sergey Levine.
\newblock {BridgeData} {V2}: {A} {Dataset} for {Robot} {Learning} at {Scale}.
\newblock In \emph{Proceedings of {The} 7th {Conference} on {Robot} {Learning}}, pages 1723--1736. PMLR, December 2023.
\newblock URL \url{https://proceedings.mlr.press/v229/walke23a.html}.

\bibitem[Wan et~al.(2025)Wan, Wang, Ai, Wen, Mao, Xie, Chen, Yu, Zhao, Yang, Zeng, Wang, Zhang, Zhou, Wang, Chen, Zhu, Zhao, Yan, Huang, Feng, Zhang, Li, Wu, Chu, Feng, Zhang, Sun, Fang, Wang, Gui, Weng, Shen, Lin, Wang, Wang, Zhou, Wang, Shen, Yu, Shi, Huang, Xu, Kou, Lv, Li, Liu, Wang, Zhang, Huang, Li, Wu, Liu, Pan, Zheng, Hong, Shi, Feng, Jiang, Han, Wu, and Liu]{wan_Wan_2025}
Team Wan, Ang Wang, Baole Ai, Bin Wen, Chaojie Mao, Chen-Wei Xie, Di~Chen, Feiwu Yu, Haiming Zhao, Jianxiao Yang, Jianyuan Zeng, Jiayu Wang, Jingfeng Zhang, Jingren Zhou, Jinkai Wang, Jixuan Chen, Kai Zhu, Kang Zhao, Keyu Yan, Lianghua Huang, Mengyang Feng, Ningyi Zhang, Pandeng Li, Pingyu Wu, Ruihang Chu, Ruili Feng, Shiwei Zhang, Siyang Sun, Tao Fang, Tianxing Wang, Tianyi Gui, Tingyu Weng, Tong Shen, Wei Lin, Wei Wang, Wei Wang, Wenmeng Zhou, Wente Wang, Wenting Shen, Wenyuan Yu, Xianzhong Shi, Xiaoming Huang, Xin Xu, Yan Kou, Yangyu Lv, Yifei Li, Yijing Liu, Yiming Wang, Yingya Zhang, Yitong Huang, Yong Li, You Wu, Yu~Liu, Yulin Pan, Yun Zheng, Yuntao Hong, Yupeng Shi, Yutong Feng, Zeyinzi Jiang, Zhen Han, Zhi-Fan Wu, and Ziyu Liu.
\newblock Wan: {Open} and {Advanced} {Large}-{Scale} {Video} {Generative} {Models}, April 2025.
\newblock URL \url{http://arxiv.org/abs/2503.20314}.
\newblock arXiv:2503.20314 [cs].

\bibitem[Wang et~al.(2025)Wang, Huang, Fang, Yang, and Ma]{wang_ATI_2025a}
Angtian Wang, Haibin Huang, Jacob~Zhiyuan Fang, Yiding Yang, and Chongyang Ma.
\newblock {ATI}: {Any} {Trajectory} {Instruction} for {Controllable} {Video} {Generation}, June 2025.
\newblock URL \url{http://arxiv.org/abs/2505.22944}.
\newblock arXiv:2505.22944 [cs].

\bibitem[Wang et~al.(2026{\natexlab{a}})Wang, Bounou, LeCun, and Ren]{wang_adajepa_2026}
Ying Wang, Oumayma Bounou, Yann LeCun, and Mengye Ren.
\newblock {AdaJEPA}: {An} {Adaptive} {Latent} {World} {Model}, June 2026{\natexlab{a}}.
\newblock URL \url{http://arxiv.org/abs/2606.32026}.
\newblock arXiv:2606.32026 [cs.LG].

\bibitem[Wang et~al.(2026{\natexlab{b}})Wang, Bounou, Zhou, Balestriero, Rudner, LeCun, and Ren]{wang_temporal_2026}
Ying Wang, Oumayma Bounou, Gaoyue Zhou, Randall Balestriero, Tim G.~J. Rudner, Yann LeCun, and Mengye Ren.
\newblock Temporal {Straightening} for {Latent} {Planning}, June 2026{\natexlab{b}}.
\newblock URL \url{http://arxiv.org/abs/2603.12231}.
\newblock arXiv:2603.12231 [cs.LG].

\bibitem[Wang et~al.(2026{\natexlab{c}})Wang, Syed, Wu, Zhang, Onol, Barreiros, Nayyeri, Dear, Zhang, and Li]{wang_Interactive_2026}
Yixuan Wang, Rhythm Syed, Fangyu Wu, Mengchao Zhang, Aykut Onol, Jose Barreiros, Hooshang Nayyeri, Tony Dear, Huan Zhang, and Yunzhu Li.
\newblock Interactive {World} {Simulator} for {Robot} {Policy} {Training} and {Evaluation}, March 2026{\natexlab{c}}.
\newblock URL \url{http://arxiv.org/abs/2603.08546}.
\newblock arXiv:2603.08546 [cs].

\bibitem[Wu and Gao(2026)]{wu_oscar_2026}
Zhuoyuan Wu and Jun Gao.
\newblock {OSCAR}: {Omni}-{Embodiment} {Action}-{Conditioned} {World} {Model} for {Robotics}, June 2026.
\newblock URL \url{http://arxiv.org/abs/2606.04463}.
\newblock arXiv:2606.04463 [cs.RO].

\bibitem[Xu et~al.(2025)Xu, Xu, Xu, Chi, Wetzstein, Veloso, and Song]{xu_Flow_2025}
Mengda Xu, Zhenjia Xu, Yinghao Xu, Cheng Chi, Gordon Wetzstein, Manuela Veloso, and Shuran Song.
\newblock Flow as the {Cross}-domain {Manipulation} {Interface}.
\newblock In \emph{Conference on {Robot} {Learning}}, January 2025.
\newblock URL \url{https://proceedings.mlr.press/v270/xu25a.html}.

\bibitem[Xu et~al.(2026)Xu, Zhang, Liu, Chen, Han, and Liu]{xu_Kinema4D_2026}
Mutian Xu, Tianbao Zhang, Tianqi Liu, Zhaoxi Chen, Xiaoguang Han, and Ziwei Liu.
\newblock {Kinema4D}: {Kinematic} {4D} {World} {Modeling} for {Spatiotemporal} {Embodied} {Simulation}, March 2026.
\newblock URL \url{http://arxiv.org/abs/2603.16669}.
\newblock arXiv:2603.16669 [cs].

\bibitem[Yang et~al.(2026)Yang, Dagli, Zook, Hadfield, Goyal, Birchfield, Ramos, and Tremblay]{yang_RoboLab_2026}
Xuning Yang, Rishit Dagli, Alex Zook, Hugo Hadfield, Ankit Goyal, Stan Birchfield, Fabio Ramos, and Jonathan Tremblay.
\newblock {RoboLab}: {A} {High}-{Fidelity} {Simulation} {Benchmark} for {Analysis} of {Task} {Generalist} {Policies}, April 2026.
\newblock URL \url{http://arxiv.org/abs/2604.09860}.
\newblock arXiv:2604.09860 [cs].

\bibitem[Yin et~al.(2024{\natexlab{a}})Yin, Gharbi, Park, Zhang, Shechtman, Durand, and Freeman]{yin_improved_2024}
Tianwei Yin, Michaël Gharbi, Taesung Park, Richard Zhang, Eli Shechtman, Fredo Durand, and William~T. Freeman.
\newblock Improved {Distribution} {Matching} {Distillation} for {Fast} {Image} {Synthesis}, May 2024{\natexlab{a}}.
\newblock URL \url{http://arxiv.org/abs/2405.14867}.
\newblock arXiv:2405.14867 [cs.CV].

\bibitem[Yin et~al.(2024{\natexlab{b}})Yin, Gharbi, Zhang, Shechtman, Durand, Freeman, and Park]{yin_Onestep_2024}
Tianwei Yin, Michaël Gharbi, Richard Zhang, Eli Shechtman, Fredo Durand, William~T. Freeman, and Taesung Park.
\newblock One-step {Diffusion} with {Distribution} {Matching} {Distillation}, October 2024{\natexlab{b}}.
\newblock URL \url{http://arxiv.org/abs/2311.18828}.
\newblock arXiv:2311.18828 [cs].

\bibitem[Yin et~al.(2025)Yin, Zhang, Zhang, Freeman, Durand, Shechtman, and Huang]{yin_Slow_2025}
Tianwei Yin, Qiang Zhang, Richard Zhang, William~T. Freeman, Fredo Durand, Eli Shechtman, and Xun Huang.
\newblock From {Slow} {Bidirectional} to {Fast} {Autoregressive} {Video} {Diffusion} {Models}, September 2025.
\newblock URL \url{http://arxiv.org/abs/2412.07772}.
\newblock arXiv:2412.07772 [cs].

\bibitem[Zhang et~al.(2018)Zhang, Isola, Efros, Shechtman, and Wang]{zhang_unreasonable_2018}
Richard Zhang, Phillip Isola, Alexei~A. Efros, Eli Shechtman, and Oliver Wang.
\newblock The {Unreasonable} {Effectiveness} of {Deep} {Features} as a {Perceptual} {Metric}.
\newblock In \emph{2018 {IEEE}/{CVF} {Conference} on {Computer} {Vision} and {Pattern} {Recognition}}, pages 586--595, June 2018.
\newblock \doi{10.1109/CVPR.2018.00068}.
\newblock URL \url{https://ieeexplore.ieee.org/abstract/document/8578166}.
\newblock ISSN: 2575-7075.

\bibitem[Zheng et~al.(2025)Zheng, Li, Wang, Liu, Kang, Feng, Zheng, Zou, Chen, Zeng, Zhang, Pang, Liu, Wang, and Zhan]{zheng_x-vla_2025}
Jinliang Zheng, Jianxiong Li, Zhihao Wang, Dongxiu Liu, Xirui Kang, Yuchun Feng, Yinan Zheng, Jiayin Zou, Yilun Chen, Jia Zeng, Ya-Qin Zhang, Jiangmiao Pang, Jingjing Liu, Tai Wang, and Xianyuan Zhan.
\newblock X-{VLA}: {Soft}-{Prompted} {Transformer} as {Scalable} {Cross}-{Embodiment} {Vision}-{Language}-{Action} {Model}, October 2025.
\newblock URL \url{http://arxiv.org/abs/2510.10274}.
\newblock arXiv:2510.10274 [cs.RO].

\bibitem[Zhou et~al.(2025)Zhou, Pan, LeCun, and Pinto]{zhou_DINOWM_2025}
Gaoyue Zhou, Hengkai Pan, Yann LeCun, and Lerrel Pinto.
\newblock {DINO}-{WM}: {World} {Models} on {Pre}-trained {Visual} {Features} enable {Zero}-shot {Planning}.
\newblock In \emph{International {Conference} on {Machine} {Learning}}, June 2025.
\newblock URL \url{https://openreview.net/forum?id=D5RNACOZEI}.

\bibitem[Zhu et~al.(2026)Zhu, Zhao, He, Su, Li, and Zhu]{zhu_Causal_2026}
Hongzhou Zhu, Min Zhao, Guande He, Hang Su, Chongxuan Li, and Jun Zhu.
\newblock Causal {Forcing}: {Autoregressive} {Diffusion} {Distillation} {Done} {Right} for {High}-{Quality} {Real}-{Time} {Interactive} {Video} {Generation}, February 2026.
\newblock URL \url{http://arxiv.org/abs/2602.02214}.
\newblock arXiv:2602.02214 [cs].

\end{thebibliography}
\endgroup

\section{Appendix}

\subsection{Additional Qualitative Evaluation}
\begin{figure*}[t!]
    \centering
    \includegraphics[width=\textwidth]{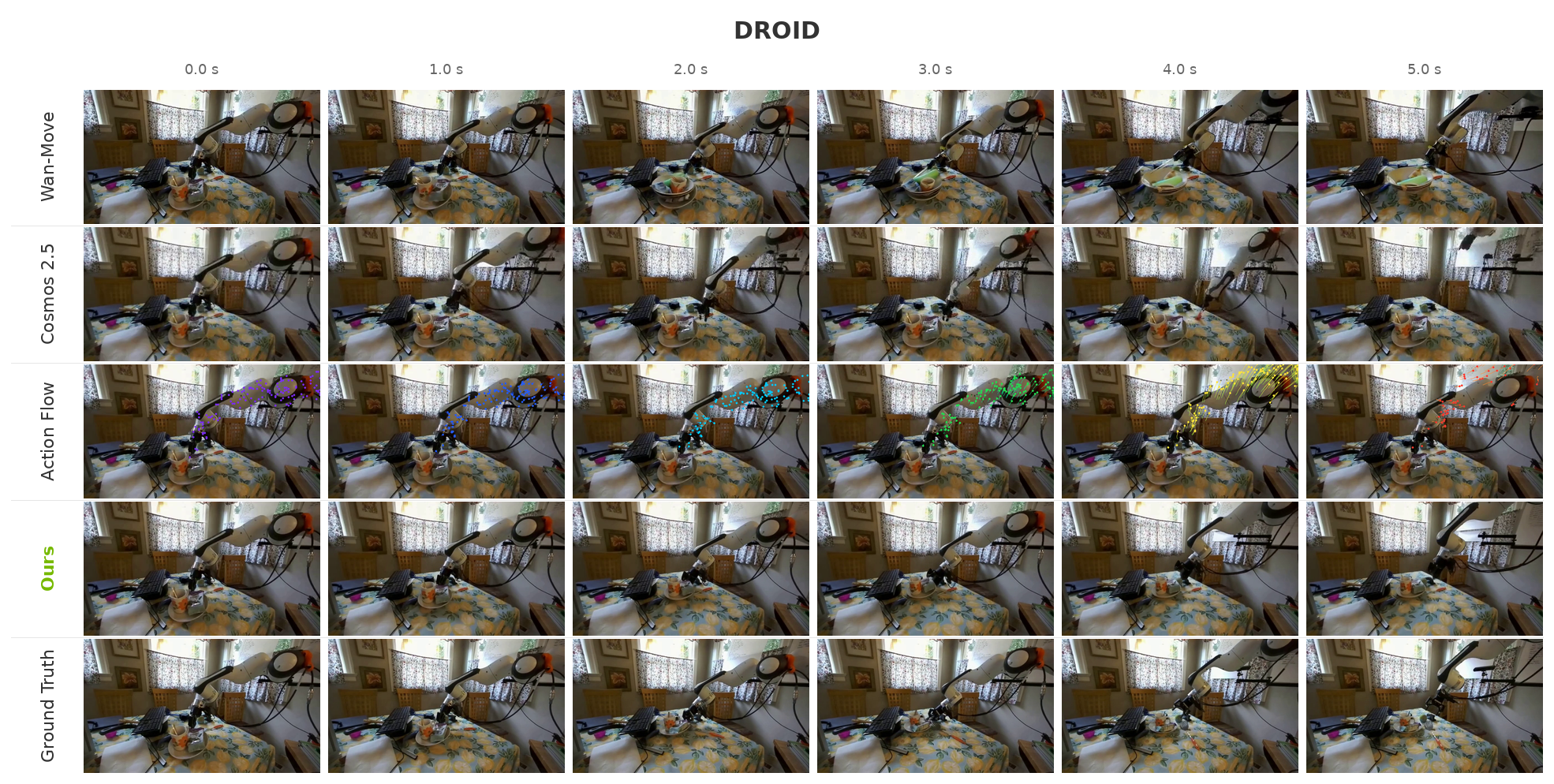}
    \caption{\textbf{Additional Qualitative Evaluation on DROID.}
    The full-size Action Flow row provides the conditioning input, while the \texttt{Ours} row shows the corresponding clean generated rollout without a flow overlay.
    All rows show the same matched validation sample at the same six synchronized timestamps.}
    \label{fig:qualitative_checkpoint_eval_droid}
\end{figure*}

Additional qualitative results on DROID are shown in \cref{fig:qualitative_checkpoint_eval_droid}.

\subsection{Action Flow Construction Details}
\label{app:flow_construction}

For metadata-rich robot data and deployment, we sample points on the visible robot surface, propagate them using robot link transforms, and project them into the camera plane as described in the main text.
At deployment, these transforms come from executing the candidate command through the robot controller and physics simulation in Isaac Lab; for recorded data, they come from observed robot states.
We retain a projected point only when it has positive camera depth, remains within the image bounds, and agrees with the rendered depth buffer within $1.2$\,cm in a $3\times3$ neighborhood around the projected pixel.
The depth buffer accounts for robot or gripper self-occlusion and includes available static scene geometry and solid-body proxies in RoboLab; unknown or unmodeled object occlusions are not explicitly removed.
This construction covers the visible embodiment, including robot arms and grippers, rather than restricting the condition to gripper motion.
For metadata-poor interaction videos, we use AllTracker to recover dense image-plane trajectories and visibility labels, then classify those tracks with grounded masks for the acting embodiment and manipulated objects.
The tracked future is used only to form training conditions; deployment flow is always derived causally from the candidate executable command.

\subsection{Training-Time Track Sampling}
\label{app:track_sampling}

We partition valid tracks into four sampling modes: \emph{none}, \emph{embodiment}, \emph{object}, and \emph{all}.
The mode selects which trajectories populate the common motion tensor and is not represented by a separate learned input token.
If a mode's required track pool is unavailable for a sample, we remove that mode and renormalize the positive probabilities of the remaining available modes; if no valid mode remains, we fall back to \emph{none}.
For the canonical I2V training mixture, the probabilities in this order are $(0.05, 0.40, 0.40, 0.15)$; dataset- and camera-specific overrides accommodate sources whose annotations or viewpoints make particular pools unavailable or unreliable.
When the corresponding pools contain enough valid tracks, the \emph{embodiment} and \emph{object} modes sample $1$--$128$ tracks, whereas the \emph{all} mode samples $256$--$1024$ tracks.
The all-mode pool is the union of embodiment tracks, object tracks, and unassigned scene tracks, from which a single subset is drawn.

\subsection{Motion-Feature Aggregation}
\label{app:motion_aggregation}

AllTracker trajectories are temporally pooled into the video's latent-frame windows: a track is visible in a window if it has any visible sample there, and its destination is the mean of its visible positions in that window.
We bilinearly sample each track's source feature from the initial-frame latent feature map.
Let $\widetilde{\mathbf{x}}_{n,k}$ and $\widetilde{m}_{n,k}$ denote a pooled track destination and its pooled visibility, and let $\widetilde{\mathbf{p}}$ denote a destination cell in the normalized latent grid.
The implemented raw Gaussian weight is exactly
\begin{equation}
    \widetilde{w}_{n,k}(\widetilde{\mathbf{p}})
    =\widetilde{m}_{n,0}\widetilde{m}_{n,k}\exp\!\left(-220\left\lVert\widetilde{\mathbf{p}}-\widetilde{\mathbf{x}}_{n,k}\right\rVert_2^2\right),
    \label{eq:raw_gaussian_weight}
\end{equation}
where $220$ is the Gaussian locality, or inverse-temperature, parameter $\beta$ from the main text.
At each destination cell, we retain the $K=2$ largest raw weights and form the propagated feature without softmax or weight-sum normalization,
\begin{equation}
    M_k(\widetilde{\mathbf{p}})=\sum_{n\in\operatorname{TopK}_2(\widetilde{\mathbf{p}},k)}\widetilde{w}_{n,k}(\widetilde{\mathbf{p}})\,\mathbf{h}_n.
    \label{eq:implemented_motion_feature}
\end{equation}
Their raw-weight sum is clamped to $[0,1]$ to form the presence gate,
\begin{equation}
    g_k(\widetilde{\mathbf{p}})=\operatorname{clip}_{[0,1]}\!\left(\sum_{n\in\operatorname{TopK}_2(\widetilde{\mathbf{p}},k)}\widetilde{w}_{n,k}(\widetilde{\mathbf{p}})\right).
    \label{eq:presence_gate}
\end{equation}

\subsection{Wan2.2 Implementation}
\label{app:wan_implementation}

The Wan2.2 I2V implementation injects the propagated features through the model's existing visual conditioning pathway rather than replacing noisy destination latents.
Concretely, its visual condition contains 16 VAE-feature channels and four mask channels: latent frame zero retains the native I2V condition, while later frames receive the raw propagated feature mass plus the pristine feature scaled by one minus the presence gate, and the gate is broadcast across the four mask channels.
The resulting visual condition is concatenated with the noisy latent at the DiT input.
For the two-denoiser I2V schedule, motion-conditioned visual features are supplied to the motion-fine-tuned denoiser; at the denoiser transition, the pristine visual condition is restored for the second, unmodified denoiser.

\subsection{Wan2.2 TI2V-5B Lightweight Variant}
\label{app:ti2v_variant}

Wan2.2 TI2V-5B provides a lightweight, single-DiT variant.
Because this model does not use the I2V visual-conditioning tensor, we form a separate 49-channel motion side input comprising 48 propagated feature channels and one presence-gate channel, concatenate it with the noisy latent at each denoising call, and widen the DiT input projection accordingly.
The motion side input remains constant across denoising steps and is neither noised nor used to re-anchor the latent state.

\subsection{Autoregressive Conversion Details}
\label{app:ar_conversion_details}

Following LongLive-2.0's direct-tuning recipe, clean-context teacher forcing forms paired clean and noisy latent streams, and a block-causal mask lets each noisy chunk attend to the preceding clean chunks and to itself while retaining the flow-matching objective in \Cref{eq:flow_matching_objective}.
Unlike earlier conversion pipelines, this direct-tuning stage does not require ODE initialization or intermediate distillation~\citep{yin_Slow_2025,zhu_Causal_2026}.
At rollout, generated chunks replace the clean history and are reused through a KV cache.
We widen the patch embedding by 49 channels to accept the 48-channel propagated feature tensor and the one-channel presence gate.
The action-flow tensor is computed once in full-window coordinates and sliced at absolute chunk offsets so trajectories are not re-anchored at chunk boundaries.

\subsection{Few-Step Distillation Details}
\label{app:few_step_distillation_details}

The student and frozen real-score model are initialized from the multi-step autoregressive checkpoint.
During on-policy backward simulation, the student generates each seven-latent-frame autoregressive chunk with four denoising steps through its causal cache before the DMD score-difference objective is applied~\citep{yin_Onestep_2024,yin_improved_2024}.
An online fake-score critic is trained on these samples, and the real--fake score difference updates the LoRA student.
We perform five critic updates per generator update and keep all three networks causal to avoid scoring samples under an incompatible bidirectional attention pattern~\citep{zhu_Causal_2026}.
Both score branches receive the same absolute-offset motion-condition slices $C_{\mathrm{motion},\mathcal{C}_j}$.
During DMD training, we set both the real- and fake-score classifier-free-guidance coefficients to zero, so neither score branch evaluates the unadapted unconditional model; the autoregressive training stage does not adapt an unconditional branch.

\subsection{VLM Judge Details}
\label{app:vlm_judge}

We use \texttt{google/gemma-4-31B-it} with deterministic decoding.
For each generated 81-frame clip, the judge receives 16 uniformly spaced frames and independently rates physical plausibility, temporal consistency, object permanence, and motion realism on a 1--5 Likert scale.
We average the four ratings for each clip and then average across clips.

\end{document}